\documentclass{article}

\usepackage{arxiv}

\usepackage[utf8]{inputenc} 
\usepackage[T1]{fontenc}    
\usepackage{hyperref}       
\usepackage{url}            
\usepackage{booktabs}       
\usepackage{amsfonts}       
\usepackage{nicefrac}       
\usepackage{microtype}      
\usepackage{lipsum}		
\usepackage{graphicx}
\usepackage{doi}

\usepackage[nohyperlinks, nolist]{acronym}
\usepackage[separate-uncertainty = true]{siunitx}
\usepackage{colortbl}
\usepackage{lscape}
\definecolor{lightgray}{rgb}{0.82, 0.82, 0.82}
\definecolor{airforceblue}{rgb}{0.36, 0.54, 0.66}
\definecolor{gainsboro}{rgb}{0.86, 0.86, 0.86}
\definecolor{lightblue}{rgb}{0.85, 0.90, 0.98}
\definecolor{lightgreen}{rgb}{0.83, 0.90, 0.83}
\definecolor{lightyellow}{rgb}{1, 0.95, 0.80}
\definecolor{lightorange}{rgb}{1, 0.90, 0.80}
\definecolor{lightviolet}{rgb}{0.88, 0.83, 0.90}

\usepackage{hhline}
\usepackage{makecell}
\usepackage{enumitem}
\usepackage{longtable}

\title{Autonomous Navigation for Robot-assisted Intraluminal and Endovascular Procedures: A Systematic Review}

\author{Ameya Pore$^{1,3, *}$, Zhen Li$^{2, 5, *}$, Diego Dall’Alba$^{1}$, Albert Hernansanz$^{3}$, Elena De Momi$^{2}$, \And Arianna Menciassi$^{6}$, Alicia Casals$^{3}$, Jenny Dankelman$^{5}$, Paolo Fiorini$^{1}$ and Emmanuel Vander Poorten$^{4}$
\\ $^{1}$Department of Computer Science, University of Verona, Verona, Italy 
\\ $^{2}$Department of Electronics, Information and Bioengineering, Politecnico di Milano, Milan, Italy
\\ $^{3}$Center of Research in Biomedical Engineering, Universitat Politècnica de Catalunya, Barcelona, Spain 
\\ $^{4}$Department of Mechanical Engineering, KU Leuven, Leuven, Belgium
\\ $^{5}$Department of Biomechanical Engineering, Delft University of Technology, Delft, Netherlands
\\$^{6}$ The BioRobotics Institute, Scuola Superiore Sant’Anna, Pisa, Italy
\\ ${}^*$Ameya Pore and Zhen Li contributed equally to this manuscript. \\
Corresponding author: Zhen Li (email: {\tt\small zhen.li@polimi.it})
}

\date{}

\renewcommand{\shorttitle}{}

\hypersetup{
pdftitle={A template for the arxiv style},
pdfsubject={q-bio.NC, q-bio.QM},
pdfauthor={David S.~Hippocampus, Elias D.~Striatum},
pdfkeywords={First keyword, Second keyword, More},
}

\begin{document}
\maketitle

\begin{abstract}
Increased demand for less invasive procedures has accelerated the adoption of Intraluminal Procedures (IP) and Endovascular Interventions (EI) performed through body lumens and vessels. As navigation through lumens and vessels is quite complex, interest grows to establish autonomous navigation techniques for IP and EI for reaching the target area. Current research efforts are directed toward increasing the Level of Autonomy (LoA) during the navigation phase. One key ingredient for autonomous navigation is Motion Planning (MP) techniques. This paper provides an overview of MP techniques categorizing them based on LoA. Our analysis investigates advances for the different clinical scenarios. Through a systematic literature analysis using the PRISMA method, the study summarizes relevant works and investigates the clinical aim, LoA, adopted MP techniques, and validation types. We identify the limitations of the corresponding MP methods and provide directions to improve the robustness of the algorithms in dynamic intraluminal environments. MP for IP and EI can be classified into four subgroups: node, sampling, optimization, and learning-based techniques, with a notable rise in learning-based approaches in recent years. One of the review's contributions is the identification of the limiting factors in IP and EI robotic systems hindering higher levels of autonomous navigation. In the future, navigation is bound to become more autonomous, placing the clinician in a supervisory position to improve control precision and reduce workload.
\end{abstract}


\section{Introduction}\label{sec:introduction}

\begin{figure}[tpb]
    \centering
    \includegraphics[width=0.7\linewidth]{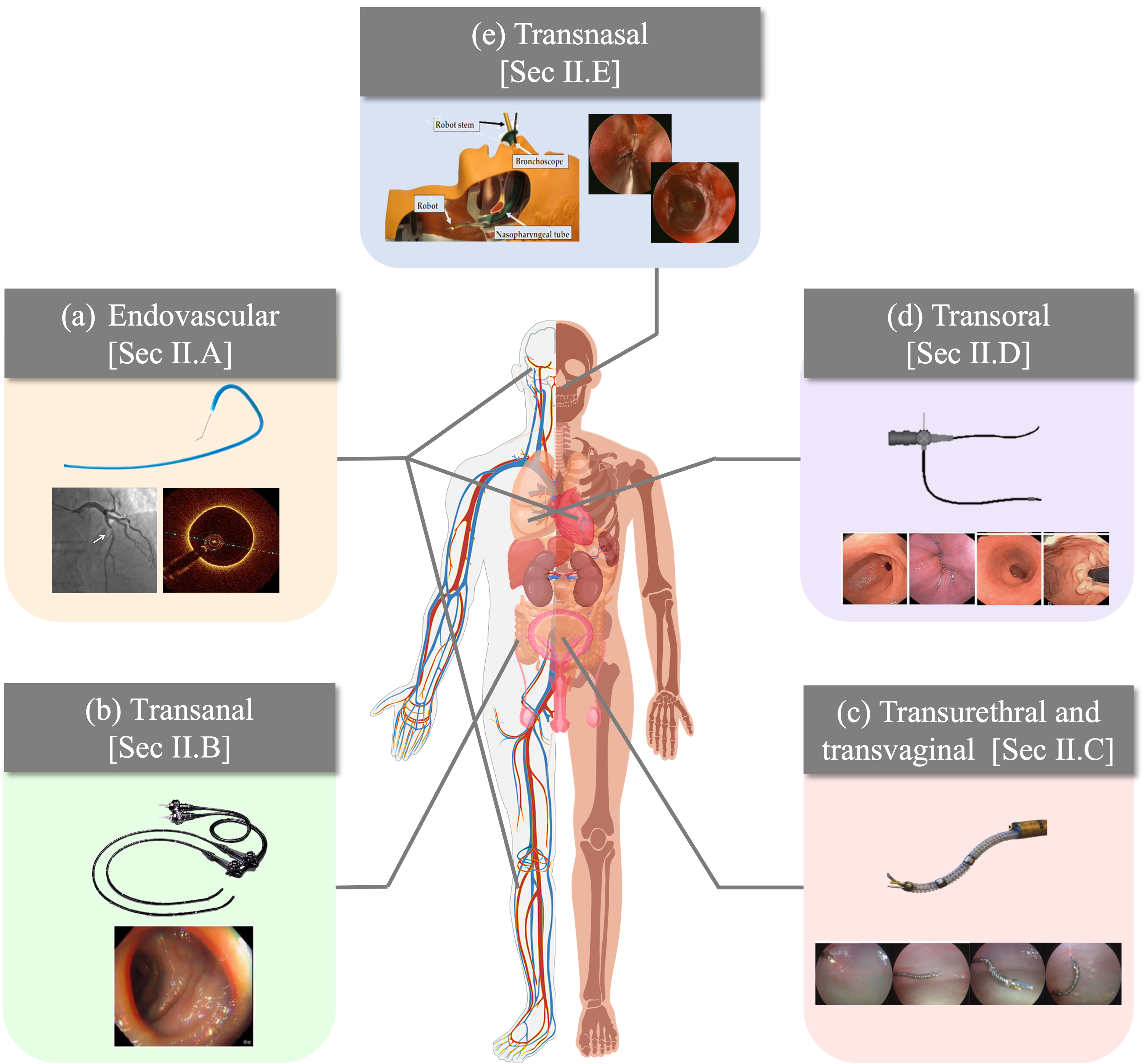}
    \caption{IPEI considered in this paper with respective standard interventional tools used and clinical target sites. (a) Endovascular catheterisation (b) Transanal colorectal procedures with a standard endoscope (c) Transurethral and transvaginal access for prostate or bladder procedures (d) Transoral procedures for airways or oesophagus (e) Transnasal procedure to access bronchi. A primary difference between IP and EI is the sensing modalities used, i.e., IP commonly use images from a camera as sensory input, whereas EI mostly use X-Ray fluoroscopy.}\vspace{-10pt}
    \label{fig:procedures}
\end{figure}

\ac{IP} and \ac{EI} are emerging medical therapies that make use of body lumens and vessels to reach otherwise difficult-to-reach regions deep into the body (Fig.~\ref{fig:procedures}). To enable these procedures, snake-like flexible instruments are needed that can adapt to the complex intraluminal and endovascular anatomy \cite{seetohul2022snake}. 
\ac{IPEI} have shown significant improvements in patient outcomes,  such as reduced blood loss, post-operative trauma, wound site infection and recovery/hospitalisation time \cite{da2020challenges}.
However, the flexible tools used in \ac{IPEI} have non-ergonomic designs. It is also difficult to control these instruments precisely as a complex mapping between input and output motion is present. This design limitation drastically increases the cognitive and physical workload of the clinician. Overall, it is well-known that clinicians undergo a long learning curve before becoming proficient in using such highly dexterous instruments~\cite{simaan2018medical}. 

IPEI are composed of several complex tasks that must be performed in the right order and following strict procedures. The first task (which may take a large proportion of time) consists of carefully navigating to reach the targeted area \cite{prendergast2018autonomous, hwang2020review}. A major challenge during this first navigation phase consists of the complexity of operating in a deformable but constrained workspace with a device that itself is quite compliant. The interventional instruments have to traverse the anatomical passageways. While doing so, they constantly keep contact with the lumen or vessels along at least a certain portion of their body length \cite{da2020challenges}. Such contacts generally happen outside the field of view due to restrictive perception of the endoluminal or endovascular tool architecture \cite{orekhov2018snake}. Contacts may be dangerous, and their response is generally hard to predict, especially when there is no direct sight of the local anatomy. 
Moreover, the movement of the tools is hard to predict. Movement at the proximal end may lead to no, limited or unexpectedly large movement of the distal tip \cite{manfredi2021endorobots}. Here, friction, slack, and deformation of the instrument and vascular or luminal wall prevent a desirable 1-to-1 relation between the proximal and distal tip motion.  

All these aspects make navigation in IPEI very challenging, and robotic systems have been introduced to improve the current situation. However, the introduction of robotic assistance has only partially reduced the procedure complexity \cite{fiorini2022concepts}, due to non-intuitive mapping between user and robot motions, limits on tool dexterity and poor shape sensing capabilities affecting situational awareness \cite{attanasio2020autonomy}. It is believed that automation could provide benefits to reduce clinicians' workload while improving the overall outcome of the procedure \cite{fiorini2022concepts, attanasio2020autonomy, haidegger2019autonomy}. 
For instance, navigation assistance could minimise path-related complications such as perforation, embolisation, and dissection caused by excessive interaction forces between interventional tools and the lumen or vessels.
Furthermore, with the increasing demand for \ac{IPEI} and the limited number of experts \cite{hargest2021five}, autonomous navigation will place clinicians in a supervisory role requiring minimal and discontinuous intervention. It will allow them to focus on high-level decisions rather than  low-level execution. 

An autonomy framework for robot-assisted \ac{MIS} was recently proposed with different Levels of Autonomy (LoA) based on robot assistance, task automation, conditional autonomy, and high level autonomy \cite{yang2017medical}.
A detailed analysis of the framework mentioned above was carried out by Haidegger \textit{et al.} and Attanasio \textit{et al.} where they map out technologies that provide distinct features at different LoA for robot-assisted \ac{MIS} \cite{haidegger2019autonomy, attanasio2020autonomy}.
These studies use a top-down approach to define LoA based on general features of robot-assisted \ac{MIS}. Hence applying these levels for specific subtasks such as navigation in \ac{IPEI} is not trivial. 
A bottom-up granular approach is required to define LoA, considering specific clinical phases. 
Therefore, this article introduces a set of characteristic features essential for defining the LoA for the IPEI navigation phase.
Characteristic features refer to subtasks associated with a specific clinical phase (i.e. IPEI navigation), such as target localisation, motion planning and motion execution.
These characteristic features are used to define the LoA for the \ac{IPEI} navigation. 
The inclusion of autonomous features raises several ethical and regulatory concerns due to incorrect robot behaviour. 
This article discusses recent regulatory developments for high-risk applications, such as autonomous robotic systems in IPEI.

One of the initial steps towards enabling autonomous navigation for \ac{IPEI} is through implementing \ac{MP} techniques \cite{patle2019review}.
\ac{MP} refers to obtaining a path from a start to a goal configuration, respecting a collision-free workspace. 

It is a well-studied problem for rigid robotic manipulators \cite{latombe2012robot}. Recent studies have explored \ac{MP} for flexible continuum robots with a large number of degrees of freedom \cite{omisore2020review,burgner2015continuum}. 
However, there is a lack of an organised survey of \ac{MP} for \ac{IPEI} and other biomedical applications using continuum robotic systems. We consider the problem of a continuum robotic system operating in a clustered and highly variable environment relevant to \ac{IPEI} scenarios.
Thus, we conduct a survey of existing \ac{MP} methods for \ac{IPEI}, the associated challenges and potential promising directions. Capsule robots are excluded from this survey since they are generally used for imaging or drug delivery with limited diagnostic capabilities. We consider IPEI robots with diagnostic capabilities, a large proportion of which are continuum robots.

The contributions of this review article are, first, to identify the \ac{LoA} for the \ac{IPEI} navigation phase; second, to provide an overview of existing MP methods that could enable autonomous navigation; and third, to provide future directions towards autonomous navigation for \ac{IPEI}.
This paper is structured as follows: Sec.~\ref{sec:pro} provides an overview of different \ac{IPEI} considered in this work, the challenges associated and the robotic systems available. Sec.~\ref{sec:autonomy} describes the LoA for \ac{IPEI} navigation and the recent regulatory measures developed. Sec.~\ref{sec:mp} introduces the survey analysis for MP methods. It presents the taxonomy and classification of MP algorithms for \ac{IPEI} procedures. Finally, the future development directions of \ac{IPEI} navigation are proposed in Sec.~\ref{sec:fd}. 
\section{Robotic automation in IPEI}\label{sec:pro}

\ac{IP} can be categorised into endoluminal and transluminal procedures \cite{vitiello2012emerging, orekhov2018snake}. Endoluminal procedures involve interventions whereby the instruments move through and stay in natural body orifices and lumens. In transluminal procedures, instruments operate in body lumens. However, they also can create incisions in lumen walls to access target sites beyond the lumen, such as natural orifice transluminal endoscopic surgery. Examples of endoluminal procedures include transoral interventions of the airways or oesophagus, transanal access to the lower digestive tract, transnasal access to bronchi and transurethral bladder and upper urinary tract procedures. Examples of transluminal procedures include transgastric and transvaginal abdominal procedures, transoesophagal thoracic and transanal mesorectal procedures (Fig.~\ref{fig:procedures}). In the context of this paper, we use \ac{IP} as an inclusive term for referring to both endoluminal and transluminal procedures. \ac{EI} use a percutaneous approach to reach target areas in the vasculature. Typical \ac{EI} include aneurysm repair, stent-graft, transcatheter aortic valve implantation, radio-frequency ablation, mitral valve repair, etc \cite{blecha2020modern}. While the technical innovation for \ac{IPEI} remains similar, \ac{EI} are carried out typically using external image guidance such as through X-Ray fluoroscopy or echography \cite{vitiello2012emerging}.

\begin{figure}[tpb]
    \centering
    \includegraphics[width=0.75\linewidth]{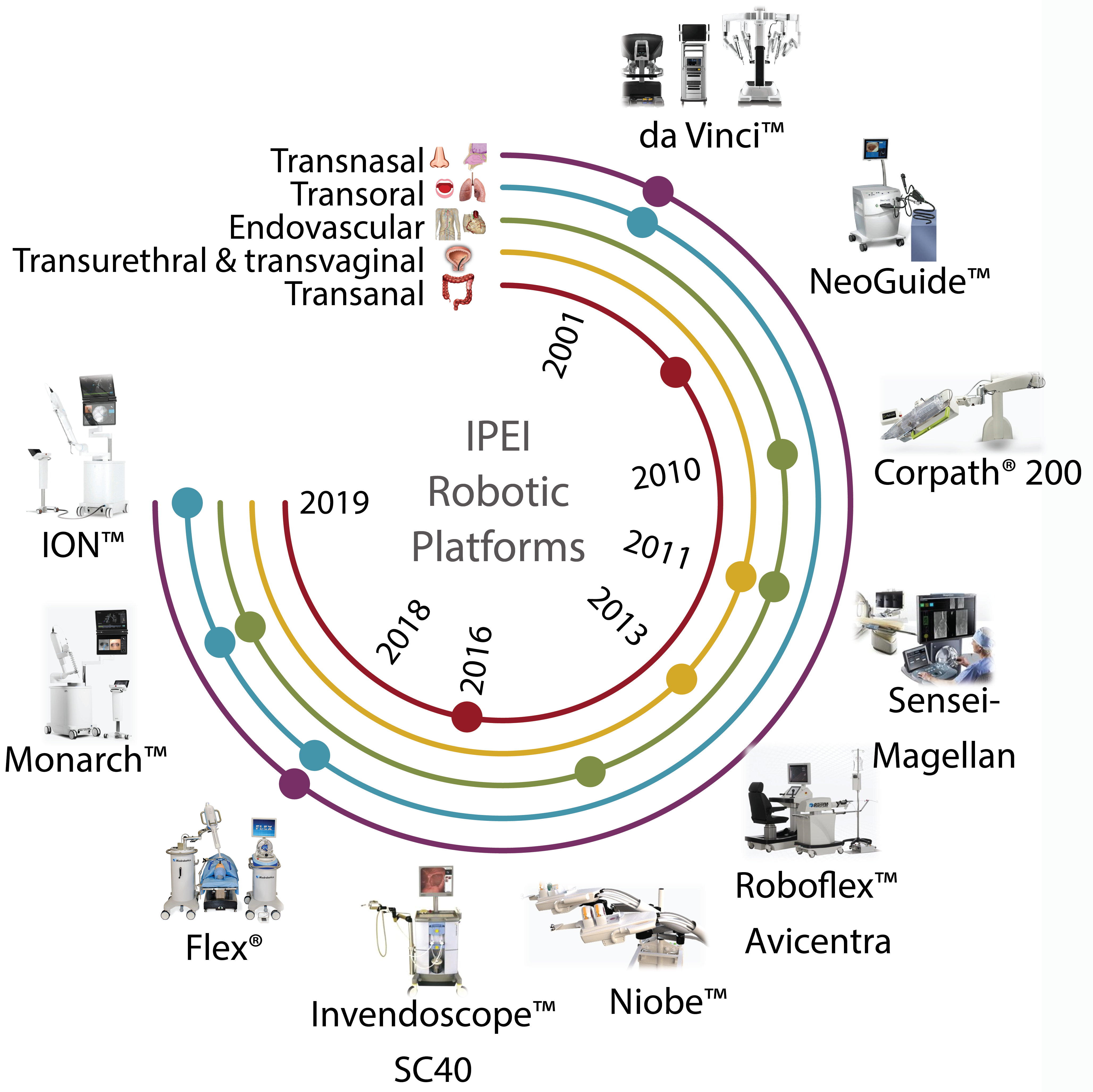}
    \caption{Selection of some commercial robotic systems for \ac{IPEI}. For endovascular interventions: Corpath™ system (Corindus, Waltham, USA) and Niobe™ system (Stereotaxis, St. Louis, USA), Sensei–Magellan (Hansen Medical, Mountain View, USA) and Monarch system (Auris Health, Redwood, USA); For transurethral and transvaginal procedures: Roboflex™ (ELMED, Ankara, Turkey) and Sensei–Magellan (Hansen Medical, Mountain View, USA); For gastrointestinal transanal procedures: Invendoscope™ (Invendo Medical, Weinheim, Germany) and Aer-O-Scope (GI View Ltd, Ramat Gan, Israel); For transnasal procedures: da Vinci (Intuitive Surgical, Sunnyvale, USA) and Flex® (Medrobotics, Raynham, USA) For bronchoscopic transoral intervention: Monarch system (Auris Health, Redwood, USA), ION™ (Intuitive Surgical, Sunnyvale, USA), da Vinci (Intuitive Surgical, Sunnyvale, USA) and Flex® (Medrobotics, Raynham, USA) are used.}\vspace{-10pt}
    \label{fig:cath}
\end{figure}

Some hospital units use consolidated robot-assisted \ac{MIS} systems \cite{villaret2017robotic} for \ac{IPEI}, however a large proportion of robotic systems consists of continuum robots \cite{orekhov2018snake,burgner2015continuum}.
Continuum robots are actuated structures that form curves with continuous tangent vectors and are considered to have an infinite number of joints and \ac{DoFs} \cite{burgner2015continuum, da2020challenges}. They have produced a step-change in medical robotics as they offer better access and safer interactions making new interventions possible. However, they are highly complex to model, sense and control \cite{da2020challenges}.
Current robotic solutions for \ac{IPEI} in the research phase are advancing the state-of-the-art through integrating new technologies that enhance the ability to recognise and interact with tissues through increased dexterity and sensory feedback \cite{attanasio2020autonomy}.
These technological advances can help in navigation guidance and building higher levels of autonomy.
Some systems are used in multiple procedures due to the lack of specific robotic technologies, multi-functionality and the ability of robotic systems to adapt to different \ac{IPEI} procedures that share similar technical or clinical characteristics \cite{peters2018review}. 
This section outlines the available robotic platforms for \ac{IPEI}. Our study considers endovascular interventions and transanal, transurethral, transvaginal, transoral, and transnasal procedures target clinical applications (See Fig.~\ref{fig:cath}).
We do not take into account procedures in which the development of continuum robotic systems is in its infancy or where the navigation phase does not constitute the predominant phase, such as auditory canal access, transvascular interventions and exploratory procedures of the lymphatic system.
\vspace{-4mm}
\subsection{Endovascular interventions}

In a general endovascular intervention, cardiologists introduce a guidewire through a small incision on the groin, the arm or the neck. The guidewire is advanced to the desired location and acts as the stable track for the catheter to follow. Two major challenges in controlling catheters and guidewires exist in this procedure. One difficulty is steering guided through a 2D fluoroscopy image \cite{rafii2014current, bonatti2014robotic}. Hence, it requires a precise understanding of the 3D anatomy projected in a 2D image plane. The other difficulty is steering the instrument tip by combining insertion, retraction and torque actions at the proximal end of the catheter and guidewire. These actions give rise to haptic feedback due to friction and collision between the catheter and the vascular walls \cite{fu2009steerable}.
Robotic advancements in computer assistance, such as enhanced instrumentation, imaging and navigation, have improved the current state of endovascular procedures. In addition, robotic platforms provide controlled steering of the catheter tip with improved stability. As a result, there is a growing interest in teleoperated robotic catheterisation systems, which offer reduced radiation exposure, increased precision, elimination of tremors and added operator comfort. 

Recent developments in CorPath™ GRX (Corindus, Waltham, USA) provide guided robotic control that allows clinicians to navigate endovascular tools through a joystick. 
Other robotic catheter systems formerly introduced by Hansen Medical (Mountain View, USA) and later acquired by J\&J robotics (New Brunswick,  USA) are the Sensei™ X and Magellan platforms. Although used for different cardiovascular applications, they are not commercially available anymore. These platforms are considered in the article since they were milestones in robotic systems for endovascular interventions \cite{peters2018review}. 
Part of this technology entered into the Monarch platform (Auris Health, Redwood, USA), which targets bronchoscopy. 
The mechanically driven Amigo™ (Catheter Robotics Inc. Budd Lake, USA) and the R-One™ (Robocath, Rouen, France) robotic assistance platform allows steering standard catheters in 3 \ac{DoFs} using an intuitive remote controller that replicates the standard handle of a catheter. The Niobe™ (Stereotaxis, St. Louis, USA) is a remote magnetic navigation system in which a magnetic field guides the catheter tip. The tip deflection is controlled by changing the orientation of outer magnets by utilising a mouse or a joystick at the workstation.
These robotic systems have reported excellent intravascular navigation. However, the absence of haptic feedback affects the procedural outcome when manoeuvring in smaller vessels like coronary, cerebral and visceral vessels \cite{pourdjabbar2017development, berczeli2021catheter}. 

\vspace{-4mm}
\subsection{Transanal IP}
Transanal colonoscopy is a widely used method for the diagnosis and treatment (screening and surveying) of colonic diseases such as \ac{CRC} \cite{ciuti2020frontiers, manfredi2021endorobots}.
In a standard colonoscopy procedure, an insertion tube is introduced through the anus and pushed forward to inspect the colonic wall \cite{yeung2019emerging}. 
Early detection and diagnosis of \ac{CRC} lesions is essential for improving the overall outcome of the patient \cite{manfredi2021endorobots,yeung2019emerging}.
The rise in the number of colonoscopies has increased the workload of endoscopists. However, not enough attention is given to the ergonomic aspects of conventional colonoscopy. Several studies have reported work-related musculoskeletal injuries of the hand, wrist, forearm and shoulder among colonoscopists \cite{you2019automatic, lemke2021colonoscopy}.
Although colonoscopy-related adverse events rarely occur, the proportion of subjects with risk factors is increasing. Severe colonoscopic complications such as perforation and bleeding can be fatal \cite{ahmed2019colonoscopy, wernli2016risks}. 
Furthermore, even well-experienced endoscopists are often limited by the lack of manoeuvrability, which can result in about 20\% of missed polyp localisation \cite{hassan2019diagnostic}. The rate of missed polyp detection varies by the polyp type, often early-stage malignancies being difficult to detect \cite{than2015diagnostic}.

Robotic Colonoscopy has been investigated to simplify the use of flexible endoscopes, reducing the procedure time and improving the overall outcome of the procedure \cite{ciuti2020frontiers}. Some cost-efficient solutions have shown advantages in reducing pain, the need for sedation, and the possibility of being disposable \cite{manfredi2021endorobots}. These platforms have a self-propelling semi-autonomous or teleoperated navigation system.
Several robotic colonoscopy platforms have received clearance to enter the market. These include the NeoGuide Endoscopy System (NeoGuide Endoscopy System Inc., Los Gatos, USA) \cite{eickhoff2006vitro}, the Invendoscope™ E210 (Invendo Medical GmbH, Weinheim, Germany), the Aer-O-Scope System (GI View Ltd., Ramat Gan, Israel)\cite{gluck2016novel}, the ColonoSight (Stryker GI Ltd., Haifa, Israel) \cite{shike2008sightline} and the Endotics System (ERA Endoscopy Srl, Pisa, Italy) \cite{tumino2017use}. 
The NeoGuide Endoscopy system and the ColonoSight are no longer commercially available. The Neoguide system is a cable-driven system that consists of 16 independent segments with 2DoFs each, position sensors at the tip to obtain the insertion depth and real-time 3D mapping of the colon. Whereas the Invendoscope™ E210 is a single-use, pressure-driven colonoscope that grows from the tip using a double layer of an inverted sleeve, reducing the forces applied to the colonic wall. The device has a working channel with electrohydraulic actuation at the tip.
The ColonoSight is composed of a reusable endoscope wrapped with a disposable sheath to prevent infection. The locomotion is provided by the air inflated inside the sleeve that covers an inner tube. The tip consists of a bendable section with two working channels. The Aer-O-Scope is a disposable self-steering and propelling endoscope that uses electro-pneumatic actuation through two sealed balloons. Recent proof-of-concept of the device showed successful caecum intubation with no need for sedation \cite{gluck2016novel}. The Endotic System uses a remotely controlled disposable colonoscope that mimics inchworm locomotion.

\vspace{-4mm}
\subsection{Transurethral and transvaginal IP}

Transurethral interventions have been used generally for bladder cancer resection, radical prostatectomy, and partial cystectomy \cite{herrell2014future}. Transvaginal access has been utilised for nephrectomy \cite{tyson2014urological}. Both these interventions use an endoscopic device to intentionally puncture a viscera (e.g. vagina, ureter and urinary bladder) to access the abdominal cavity and perform intra-abdominal operations \cite{bazzi2012natural}.
There are considerable challenges that limit the widespread adoption of transurethral and transvaginal access for urological applications, such as the unmet need for dedicated specially designed instruments resulting in lack of distal dexterity, limited tool accuracy, and limited depth perception \cite{chen2020review, tyson2014urological}. 
These factors lead to the under-resection of tumours and difficulty in enucleating tissue with minimal tilting of the rigid tools and the urethral anatomy, motivating research in robot-assisted techniques \cite{herrell2014future}.

In 2008, \ac{fURS} was accomplished using the Sensei–Magellan system (Hansen Medical, Mountain View, USA), which was designed for cardiology and angiography \cite{rassweiler2018robot}. 
Since 2010, ELMED (Ankara, Turkey) developed the Roboflex™ Avicenna for \ac{fURS} that directly drives the endoscope and an arm enabling rotation by a joystick. Compared to traditional flexible ureteroscopy, this system's advantage lies in improved movement precision and better ergonomics \cite{gandaglia2016novel}.

\vspace{-3mm}
\subsection{Transoral IP}

Conventional \ac{TOE} is the standard diagnostic method used to examine the oesophagus, stomach, and proximal duodenum. In \ac{TOE}, varying lengths of flexible endoscopes are used, e.g. gastroscopes (925mm–1.1 m), Duodenoscopes (approximately 1.25 m) and Enteroscopes (1.52- 2.2 m) \cite{valdastri2012advanced}.
The diagnostic and therapeutic capabilities of \ac{TOE} strongly correlate with the technical and decision-making skills of the operator with a steep learning curve \cite{marlicz2020frontiers}. 
Standard endoscopic surgical approach for laryngeal lesions uses laryngoscope, microscope and laser \cite{de2020trans}. This approach requires the surgeon to work within the limits of the laryngoscope and gain line-of-sight observation to complete the operation \cite{de2020trans}. Transoral access is also used for bronchoscopy to reach the lungs farther down the airways. Conventionally, a bronchoscope is used for such procedures \cite{prakash1994bronchoscopy}. However, the average diagnostic yield remains low because of limited local view in the peripheral airways \cite{kennedy2020computer}. Electromagnetic navigation was introduced to guide the bronchoscope through the peripheral pulmonary lesions, but it lacked direct visualisation of the airways, hence motivating the need for robotic assistance \cite{agrawal2020robotic}.

Available robotic systems for \ac{TOE} includes the EASE system (EndoMaster Pte, Singapore) and EndoSamurai™ (Olympus Medical Systems Corp., Tokyo, Japan). The EASE system is based on a teleoperated device that remotely controls the endoscopic medical arms. The EndoSamurai™ system consists of instruments mounted at the end of the endoscope for submucosal dissection procedures. Some other robotic systems in an early development phase are reviewed in \cite{marlicz2020frontiers}. 

Commercially available systems for laryngeal procedures are the da Vinci Robotic System (Intuitive Surgical, Sunnyvale, USA) and the Flex Robotic System (Medrobotics, Raynham, USA) \cite{villaret2017robotic}. The Flex robotic system includes a rigid endoscope controlled through a computer interface, with two external channels for flexible instruments.

In robotic bronchoscopy, Monarch™ (Auris Health Inc, Redwood, USA) is pioneering robotic endoscopy. The platform consists of an outer sheath, an inner bronchoscope with 4DoF steering control, electromagnetic navigation guidance and continuous peripheral visualisation \cite{agrawal2020robotic}. 
Another robotic platform called ION™ Endoluminal System by Intuitive Surgical includes an articulated, flexible catheter with shape sensing capabilities, which provides positional and shape feedback along with a video probe for live visualisation while driving the catheter. \cite{agrawal2020robotic}.
\vspace{-3mm}
\subsection{Transnasal IP}

Systems for the transnasal procedure have been investigated with several exploratories in mind. These procedures, ranked according to the distance to the target from the entry point include transnasal navigation for sinuses, transnasal skull base procedure, and transnasal micro-procedure of the upper airways.

One of the challenges with diseases of the sinuses lies in the difficulty of monitoring their progression, obtaining a biopsy, and facilitating intervention in the frontal and maxillary sinuses while avoiding visible scarring or obliteration of bone scaffolds of the nose. Conventionally, a flexible endoscope is used in clinical practice \cite{stammberger1990functional}. Skull base surgeries are carried out through transnasal access. A typical target for these surgeries is the removal of pituitary gland tumours through a transsphenoidal approach \cite{burgner2013telerobotic, madoglio2020robotics}. The standard endoscopic approach for these surgeries is limited by restricted access, cumbersome manual manipulation of interventional tools near susceptible anatomy and lack of distal dexterity \cite{tanuma2016current}.

Another interventional target using transnasal access is the upper airways and throat \cite{simaan2009design}.
\ac{TNE} is performed using an ultrathin endoscope with a shaft diameter of \SI{6}{\mm} which is inserted through the nasal passage. Once the instrument is beyond the upper oesophageal sphincter, endoscopy is conducted in the standard fashion. However, there are some technical limitations of \ac{TNE}, namely, a smaller working channel can result in limited suction and the availability of fewer endoscopic accessories.

In general practice, the robotic systems mentioned in transoral approaches such as da Vinci Robotic System (Intuitive Surgical, Sunnyvale, USA) and Flex® Robotic System (Medrobotics Corp., Raynham,  USA) are also used in transnasal interventions \cite{villaret2017robotic}. The Flex® Robotic System is an operator-controlled flexible endoscope system primarily designed for an Ear-Nose-Throat procedure that includes a steerable endoscope and computer-assisted controllers, with two external channels for the use of compatible \SI{3.5}{\mm} flexible instruments. However, specific robotic systems with appropriate ergonomics and dimensions suited for transnasal passage are still under development \cite{villaret2017robotic}.
\begin{footnotesize}
\begin{table*}
\centering
\caption{Descriptive classification of LoA for IPEI navigation. H: Performed by a human operator, M: Performed by a machine. H/M: Performed by a human, assisted by a machine, M/H: Performed by a machine, assisted by a human. M\textsuperscript{1}: Performed under human supervision.} \label{tab:autonomy_table} 
\begin{tabular}{p{0.5cm}p{9.0cm}p{1.5cm}p{1.2cm}p{1.5cm}}
\hhline{|=|=|=|=|=|}
\textbf{LoA} & \textbf{Description} & \makecell[l]{\textbf{Target} \\ \textbf{localisation}} & \makecell[l]{\textbf{Motion} \\ \textbf{planning}} & \makecell[l]{\textbf{Execution \&} \\ \textbf{re-planning}}\\ \toprule
0 & \textit{Direct robot control}: The clinician exclusively controls all cognitive functions without any support or assistance \cite{yang2017medical}. Most IPEI systems used in clinical practice operate at Level-0 autonomy.
& H & H & H \\
\midrule
1 & \textit{Navigation assistant}: The human operator maintains continuous control of the robotic navigation intraoperatively; however, it is assisted robotically during the execution of the motion. Other cognitive functions are carried out manually.  
& H & H or M & H/M \\
\midrule
2 & \textit{Navigation using waypoints}: The operator provides discrete high-level navigation tasks such as waypoints or predefined trajectories. These trajectories are derived during preoperative planning. The robot carries out the required motion between the waypoints during the execution time, with the clinician in a supervisory role to approve or override the strategy. 
& H & M/H & M\textsuperscript{1} or M/H \\
\midrule
3 & \textit{Semi-autonomous navigation}: The final goal of navigation is provided by a human operator, and
the system generates the strategies required to carry out the complete navigation task.
During the execution time, it relies on the operator's supervision to approve or override the choice. 
In IPEI navigation, the robot would extract waypoints and then plan the trajectory to reach the point. 
& H & M & M\textsuperscript{1} \\
\midrule
4 & \textit{High-level autonomous navigation}: This level is characterised by the ability of the system to make clinical decisions and execute the control solution under the clinician's supervision. The system should interpret preoperative imaging modalities such as CT, MRI and ultrasound to detect target regions and extract all the information required for proper navigation.   
& M & M & M\textsuperscript{1}\\
\hhline{|=|=|=|=|=|}
\end{tabular}
\end{table*}
\end{footnotesize}
\section{Levels of Autonomy (LoA)}\label{sec:autonomy}
One of the promising features of upcoming IPEI robotic systems is autonomy since it provides the ability to perceive, analyse, plan and take actions automatically \cite{nof2009automation}. An autonomous robotic system can deal with non-programmed situations and has the capability of self-management and self-guidance \cite{olszewska2017ontology}. The most notable aspect of autonomy is the transfer of decision-making from a human operator to a robotic system. 
To allow this transfer, two conditions must be met \cite{chen2021automation}. 
First, the operator must transfer the control to the robotic system, including the related responsibilities (i.e., the human operator must ``trust" the autonomous system).
Second, the system must be certified, i.e., it must fulfil all ethical, legal and certification requirements. 
However, these certification standards are not fully developed for medical robotic systems due to a lack of consideration, and clear understanding of autonomy \cite{fisher2021towards}. 
Therefore, we first introduce the ethical and regulatory aspects related to autonomy in Sec.~\ref{sec: ethical}, then we define generic LoA in Sec.~\ref{sec:definition}, while in Sec.~\ref{sec: autonomyipei}, we present the specific LoA for IPEI navigation systems.

\vspace{-2mm}
\subsection{Ethical and regulatory aspects of autonomy} \label{sec: ethical}
The ethical concerns can be addressed from multiple perspectives, including human rights, law, economics, policy and ethics \cite{ethics}. We highlight the viewpoints of medical robot practitioners. When transferring the decisions from a human operator to an autonomous system, one of the main ethical concerns is the consequences of errors resulting from the decisions taken. These errors can be due to incorrect robot behaviours, leading to hazardous situations \cite{o2019legal}.
Hence, robot-assisted intervention is considered a high-risk category \cite{aiact}. To address the ethical concerns, the European Commission proposed a regulatory framework for AI applications in the high-risk category, known as the Artificial Intelligence Act (AI Act) \cite{aiact}. Similar efforts are being formalised in the United States under the National AI Initiative \cite{nationalaiact}. The AI Act describes the role of a human operator: the obligation to provide human supervision, the right for a human to override an automated decision, and the right to obtain human intervention which forbids full autonomy. Therefore, human intervention needs to be carefully designed into the system at different levels of integration \cite{fiazza2021icar}. Furthermore, the reliability of medical robotics is associated with the notion of certification, which requires legal approval that the system has reached a particular standard. Several regulatory standards exist in the robotics domain; for instance, the standards for medical electrical systems are defined by the International Electrotechnical Commission (IEC) in the Technical Report (TR) 60601-4-1 \cite{medicaldevices}, which guides risk management, basic safety and essential performance towards systems with some degrees of autonomy. These regulatory standards are not fully developed for robot-assisted intervention, and the introduction of LoA could support this development by facilitating the system verification and validation with improved risk management \cite{fisher2021towards}. As a consequence of upcoming regulations like the AI Act, it is expected that earlier phases of the design process will progressively consider safety and system integration concerns.

\vspace{-2mm}
\subsection{Levels of Autonomy: Definition} \label{sec:definition}

Quantifying system autonomy based on its capabilities presents a significant challenge due to different levels of advances in underlying technologies. 
In medical robotics, as the autonomous capabilities of the robot are increasing, the role of medical specialist 
is shifting from manual dexterity and interventional skills towards diagnosing and high-level decision-making.

Prior work identified five LoA for medical robot systems considering a complete clinical procedure and the capability of a human operator/clinician \cite{yang2017medical, haidegger2019autonomy}. At level 0, the robot has no decision autonomy, and the clinician controls all aspects of the system. 
i.e., the clinician exclusively controls it. At level 1, the robot can assist the clinician, while at level 2, it can autonomously perform an interventional subtask. At level 3, the robot can autonomously perform longer segments of the clinical procedure while making low-level cognitive decisions. Finally, at level 4, the robotic system executes the complete procedure based on human-approved clinical plans or surgical workflow. Few studies have defined level 5, which refers to full autonomy in which the robotic clinician can perform the entire procedure better than the human operator; hence human approval is not required \cite{yang2017medical, haidegger2019autonomy}.
However, level 5 is still in the realm of science fiction, so we consider it outside the scope of this article. 
In higher LoA, the robot responding to various sensory data will be highly sophisticated while it could replicate the sensorimotor skills of an expert clinician more closely.  

Attanasio \textit{et al.}  \cite{attanasio2020autonomy} outlined the enabling technologies and the practical applications for different levels. Haidegger \textit{et al.} provided a top-down classification of LoA for general robot-assisted \ac{MIS} \cite{haidegger2019autonomy}. Their classification considers four robot cognitive functions (i.e., generate, execute, select and monitor options), where the overall LoA is the normed sum of the four system functions assessed on a linear scale, ``0" meaning fully manual and ``1" fully autonomous. 

In clinical practice, an interventional procedure workflow is decomposed into several granular levels, such as phase, steps, and gestures \cite{katic2015lapontospm}. Many of the interventional phases and skills that are used in robot-assisted \ac{MIS} are not considered in IPEI, e.g., luminal navigation. Hence, LoA defined for robot-assisted \ac{MIS} can not be directly applied for IPEI.
Moreover, using the proposition provided by Haidegger \textit{et al.}, it is challenging to identify a clear boundary between human and automated control required for specific phases/steps of robot-assisted \ac{MIS}. It introduces an additional problem of defining the system's overall level that implements different LoA for different phases of the procedure. Hence, we propose a bottom-up solution where an intermediate LoA is defined for specific interventional phases. Having knowledge of a subtask will enable a better understanding of the amount of human intervention required at a granular scale. A bottom-up classification would better estimate the overall system autonomy since underlying phases can be at a different intermediate LoA. 
Moreover, it can be applied to all medical procedures, from robot-assisted \ac{MIS} to IPEI. The target of this article is IPEI navigation; hence we define the intermediate LoA for this interventional phase. 

\vspace{-3mm}
\subsection{LoA for \ac{IPEI}} \label{sec: autonomyipei}

LoA for robot-assisted \ac{MIS} has been derived from the degree of autonomy introduced by ISO, who, jointly with IEC, created a technical report (IEC/TR 60601-4-1) \cite{medicaldevices} to propose an initial standardisation of autonomy levels in medical robotics. The report parameterises \ac{DoFs} along a system's four cognition-related functions: generate, execute, monitor and select options strategy. A similar classification approach has been followed by Haidegger \textit{et al.} for robot-assisted \ac{MIS}. We identify three specific cognitive functions for an IPEI navigation task: 1) Target localisation, 2) Motion planning, and 3) Execution and replanning. Target localisation is usually based on preoperative images, such as \ac{CT}, \ac{MRI} or X-Ray imaging. It is a critical feature, as inaccurate target identification can lead to inaccuracies in the subsequent steps. MP can be considered in two phases: preoperative and intraoperative. Preoperative MP refers to the planning performed before the procedure based on multi-modal medical images \cite{ravigopal2022fluoroscopic}. This may be done in static virtual models of the lumen or vessels. Execution and replanning is an intraoperative phase to carry out the required motion to reach the target while continuously replanning intraoperatively. It can include target relocalisation when adjustment is needed due to unexpected situations.

\begin{figure}[tb]
    \centering
    \includegraphics[width=0.75\linewidth]{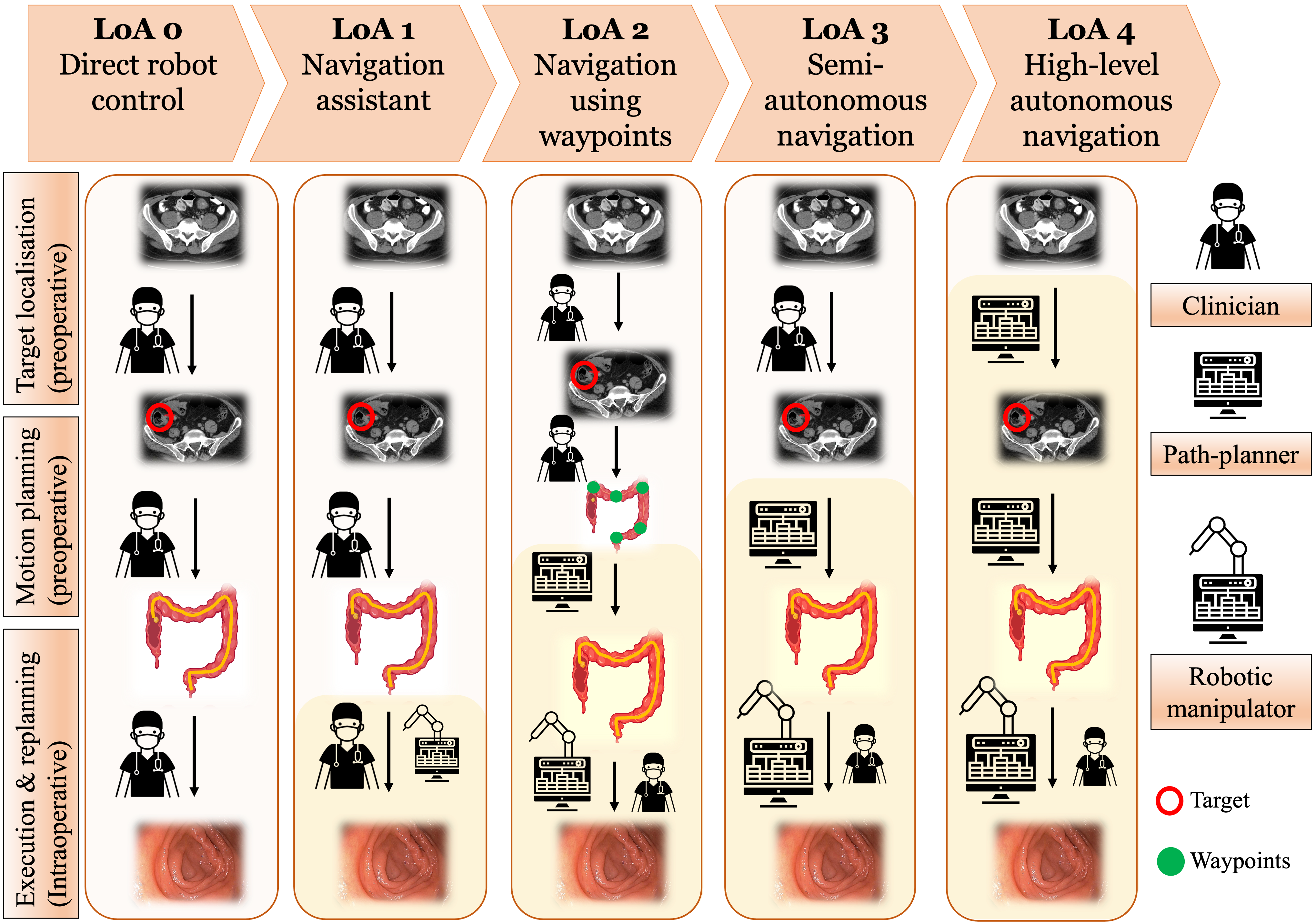}
    \caption{Case study of LoA for endoscopic navigation for transanal IP. 
    The complete navigation task is divided into three cognitive functions: target localisation through preoperative imaging, planning the motion preoperatively and executing the motion. (Row 1) Target localisation using preoperative images: The identified target is depicted with a red circle.
    (Row 2) Preoperative MP: Path representation inside the colon shown with a yellow line.
    (Row 3) Intraoperative motion execution and MP: Intraoperative endoscopic visualisation.
    (left to right). LoA0-LoA4 respectively. For each level, we indicate the agent that operates each cognitive function. Agent refers to either a human operator, path-planning system or robotic manipulator. In the case of two agents, the supervisor agent is depicted on the right side, while the main agent executing the actions is on the left and its icon is larger.
    }
    \label{fig:autonomy_levels}
\end{figure}

Table.~\ref{tab:autonomy_table} illustrates the LoAs defined for \ac{IPEI} navigation. In LoA 0, all the features from target localisation, MP and motion execution are carried out by a human operator. Commercially available robotic system (as described in Sec. II) can be considered in this category since the human operator has complete control of the robotic motion.
LoA 1 is characterised by target localisation and preoperative planning manually carried out by the clinician. The clinician executes the actual motion with the assistance of the robotic system. Systems that use external tracking devices and registration methods to align the preoperative data with the intraoperative condition and support the clinician in executing a clinical procedure can be considered LoA~1 \cite{khare2015hands, zheng20183d, zang2019optimal}. Taddese \textit{et al.} developed a teleoperated magnetically controlled endoscope, where the system provides navigation assistance by controlling the magnetic field \cite{taddese2018enhanced}. These systems represent the first implementations of LoA~1, where the manipulator executes the command imparted by the operator. In LoA 2, the robotic system fully controls the specific navigation steps. Target localisation is carried out by the clinician, who provides input in the form of waypoints or demonstration trajectories. The path planner uses these cues to generate a global trajectory. Further, the robotic system carries out the required motion indicated by the path planner. During execution, the human operator supervises the autonomous navigation and approves the robot's actions or overrides it (to comply with AI Act indications). In LoA~3, after target localisation by the clinician, the path planner generates the global path in the preoperative phase without any manual intervention. This level includes automatically splitting the entire navigation task into specific subtasks that could be performed autonomously. The robotic system executes the motion indicated by the path planner and adapts to environmental changes through real-time replanning. The local real-time knowledge will provide information regarding the anatomical environment, and the motion will be adjusted as the autonomously steering is performed. All the features from target localisation, MP and execution are autonomously carried out without any human intervention by a system reaching LoA 4. The main difference between LoA~3 and LoA 4 is the addition of automatic target identification. This additional feature requires enabling technologies such as autonomous segmentation of organs to detect abnormal tissues such as polyps, automatic localisation and shape sensing mechanisms \cite{fiorini2022concepts} (Sec.~\ref{subsec:rob_cap}). Fig.~\ref{fig:autonomy_levels} shows a case study of LoA for the transanal IP. In the next section, the proposed LoA will be used to classify all the work considered in the field of IPEI navigation.

\section{Systematic review of MP for IPEI }\label{sec:mp}
\subsection{Literature review}\label{sec:lr}

A systematic analysis was conducted, following the PRISMA methodology \cite{page2021prisma}, to survey the developments of automation and MP in \ac{IPEI}. 

\subsubsection{Search method}\label{sec:search}

A systematic analysis was conducted using the following digital libraries: \texttt{Google Scholar}, \texttt{Scopus} and \texttt{IEEE Xplore}. Search queries were programmatically generated from the search term matrix. Query results were automatically retrieved and checked for duplicates via the \texttt{Scopus} API. The list of references was saved as a \textit{.csv} file and manually evaluated according to the inclusion criteria. All items that did not meet the inclusion criteria were excluded. The search terms used in this survey were chosen by generalising the term ``motion planning for intervention". 

\begin{figure}[tb]
    \centering
    \includegraphics[width=0.9\linewidth]{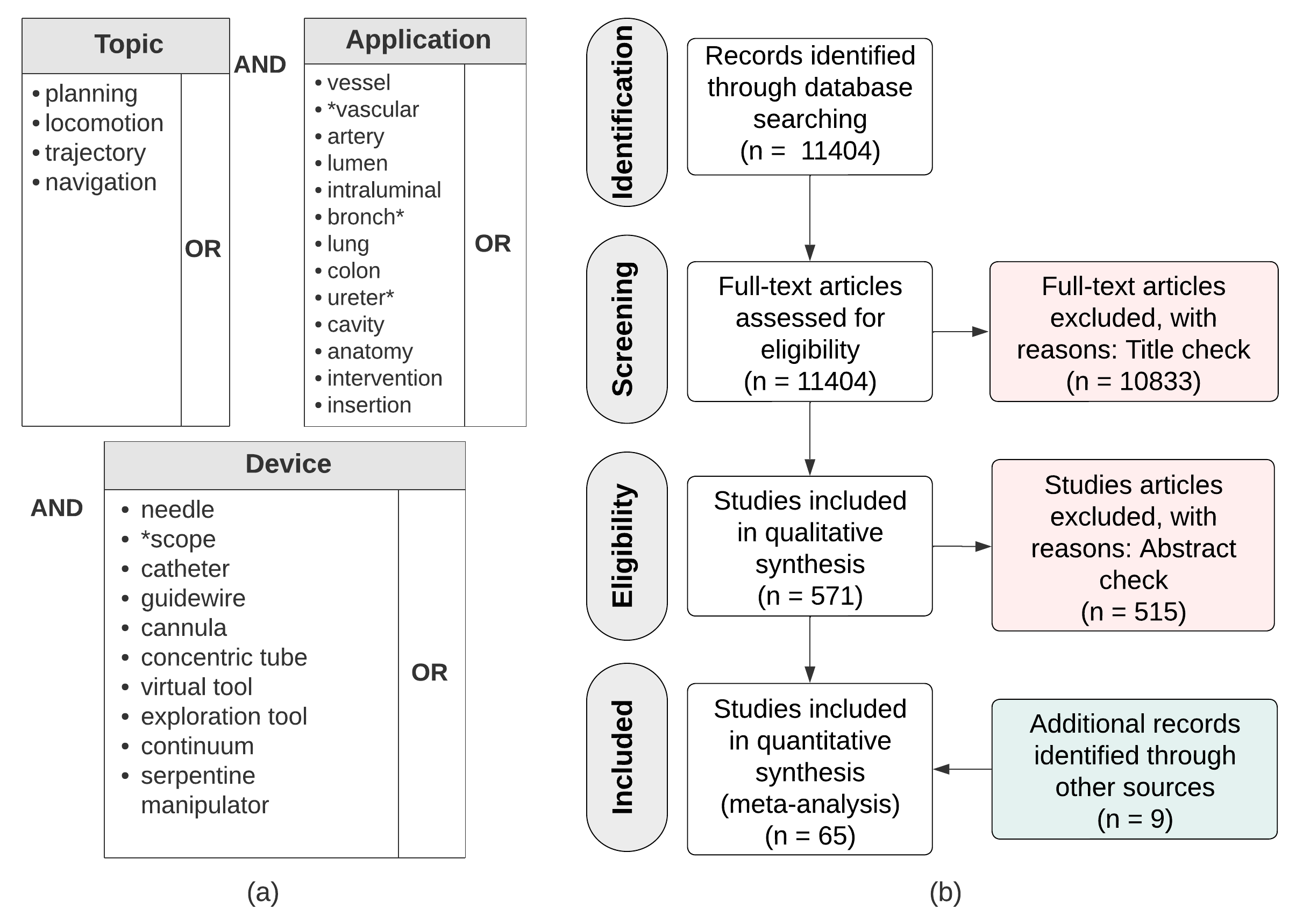}
    \caption{(a) Search matrix used for the survey (b) Prisma flow diagram summarizing how the systematic review was conducted.}
    \label{fig:meta}
\end{figure}

Search terms are combined with the logical operators AND and OR such that a large search space can be covered in sufficient detail. Fig.~\ref{fig:meta} provides an overview of all the search terms and the flow of the conducted review. This matrix, once all possible combinations have been exhausted, yields 520 entries. 
To automatically manage all the generated entries and remove the duplicates, a python library, pybliometrics, was used \cite{rose2019pybliometrics}. 
The cutoff date for the earliest work included is 2005, and the latest work is from July 2022.

\subsubsection{Selection criteria}\label{sec:selection}

The paper was selected by:
\begin{enumerate}[label=(\roman*)]
\item considering only continuum robots (excluding capsule mobile robots \cite{vuik2021colon, meng2019motion}) for \ac{IPEI};
\item excluding low-level controller studies based on force control, position control, impedance control and similar;
\item considering only full papers drafted in English. Extended abstracts reporting preliminary findings were omitted;
\item excluding transluminal procedures that require incisions such as hydrocephalus ventricles.
\end{enumerate}

\subsubsection{Post processing and analysis}\label{sec:postanalysis} 

The search script returned 11404 references. Prisma flow diagram in Fig.~\ref{fig:meta}b summarises how the systematic review was conducted. 10833 references and 515 references were excluded after title check and abstract check, respectively. Additional 9 references were included manually because search results did not cover 100\% of the current studies for different reasons. Finally, this process yielded a list of 65 references.

\begin{figure}[tb]
    \centering
    \includegraphics[width=0.7\linewidth]{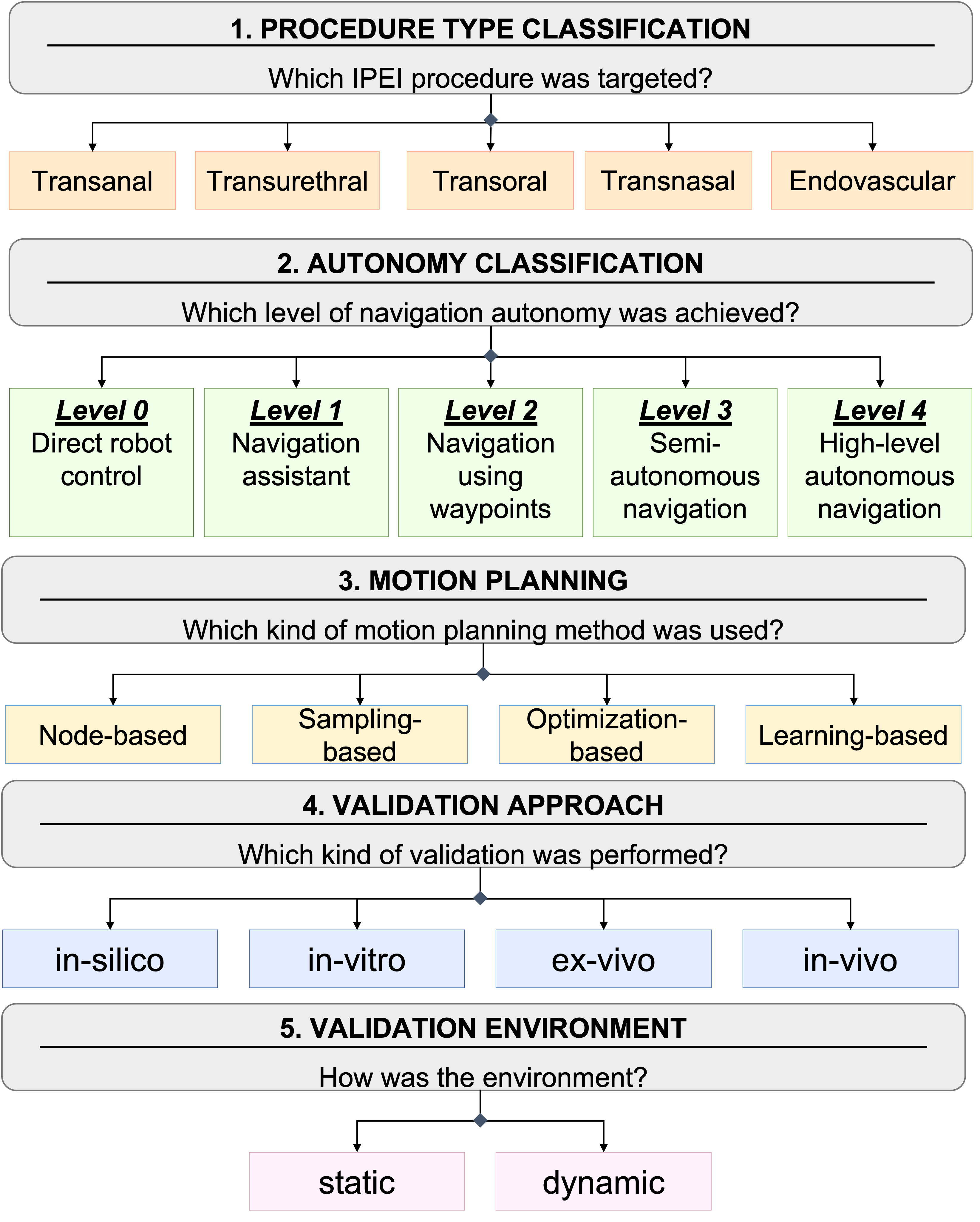}
    \caption{Schematics of the analysis carried out for each paper. These criteria include the targeted procedure, the LoA, the MP method, the validation and the dynamics of the environment.}
    \label{fig:clas}
\end{figure}

The outcomes of various studies were classified based on several criteria, shown in Fig.~\ref{fig:clas}, including the targeted procedure, the LoA, the MP method, the validation, and the environment's dynamics. The MP methods are categorised into subgroups presented in Fig.~\ref{fig:taxonomy} for an in-depth analysis. The summary of the state-of-the-art on \ac{IPEI} MP publications are presented in Table~\ref{tab:reflist}, and its development is shown in Fig.~\ref{fig:pub}a. Besides the MP approach, we have highlighted the distribution of the targeted \ac{IPEI} procedures in Fig.~\ref{fig:pub}b. Moreover, Table~\ref{tab:reflist} shows that some studies involved intraoperative path replanning with a dynamic environment (last column).

\subsection{Taxonomy on MP for \ac{IPEI} navigation}\label{sec:plan}

MP has been a well-documented field for navigation tasks since the 1980s, supporting robotic manipulators and mobile platform operations in indoor and outdoor industrial applications. During \ac{MP}, robot characteristics are usually considered to find a feasible path solution, such as its geometrical dimensions to avoid collisions and kinematic constraints to respect its movement capability. The robot kinematics describes the relationship between the configuration and task spaces \cite{siciliano2008springer}. The configuration space $\mathcal{C}$ is defined as all robot configurations. The task space $\mathcal{T}$ is referred to as the workspace that the robot can reach for each specific configuration $\mathbf{q}$. The robot kinematics can be expressed in a general form as
\begin{equation}
    \mathcal{T} = f(\mathbf{q}) \quad \mathbf{q} \in \mathcal{C}
\end{equation}

MP is an essential component of autonomous IPEI robotic systems, even under complex operating conditions and stringent safety constraints. As shown in Fig.~\ref{fig:taxonomy}, MP methods can be decomposed into four sub-groups by adjusting the taxonomy of path planning in general robots from \cite{yang2016survey}: node, sampling, optimisation and learning-based techniques. The node-based (or graph-based) algorithms use a graph-searching strategy along with a tree structure. The sampling-based algorithms construct a tree structure based on random samples in a configuration space. Therefore these methods find a collision-free path and ensure compatibility with the robot's motion capabilities. Optimisation-based algorithms formulate the MP problem as a mathematical problem by minimising or maximising an objective function with respect to some constraints and obtaining the optimal case through a solver. 
Learning-based methods use a Markov decision process to learn a goal-directed policy based on a reward function.
A brief general definition of different MP methods is provided in Table.~\ref{tab:env}, while Table III summarizes different MP works for IPEI applications.

\begingroup
\small
\begin{longtable}{ p{0.5cm} | p{2.5cm} | p{11.0cm}} 
\caption{Background of path-planning methods.} \label{tab:env} \\
\textbf{No.} & \textbf{Path Planning} & \textbf{Description} \\ \toprule
1. & \textbf{Node-based} &  \\ \midrule
a. &  Centerline-based Structure (CBS) & This method is long-established to keep the tip of the instruments away from the walls \cite{cheng2012enhanced}. A tree structure is built from the anatomical information of the lumen, where each node contains the information of the lumen centerline position and the corresponding lumen radius. \\
b. & Depth First Search (DFS) & DFS algorithm traverses a graph by exploring as far as possible along each branch before backtracking \cite{tarjan1972depth} \\
c. &  Breadth First Search (BFS) & BFS algorithm \cite{dechter1985generalized} starts at the tree root and explores the k-nearest neighbor nodes at the present depth before moving on to the nodes at the next depth level. \\ 
d. &  Dijkstra & The Dijkstra algorithm~\cite{dijkstra1959note} is an algorithm for finding the shortest paths between nodes in a graph. It is also called Shortest Path First (SPF) algorithm. The Dijkstra algorithm explores a graph by expanding the node with minimal cost. \\
e. &  Potential field & Artificial potential field algorithms \cite{hwang1992potential} define a potential field in free space and treat the robot as a particle that reacts to forces due to these fields. The potential function is composed of an attractive and repulsive force, representing the different influences from the target and obstacles, respectively. \\
f. &  A* \& Lifelong Planning A* (LPA*) &  A-star \cite{hart1968formal} is an extension of the Dijkstra algorithm, which reduces the total number of states by introducing heuristic information that estimates the cost from the current state to the goal state. \\
g. &  Wall-following & Wall-following algorithms move parallel and keep a certain distance from the wall according to the feedback received from sensors. \\
\midrule
2. &  \textbf{Sampling-based} &  \\ \midrule
a. & Rapidly-exploring Random Tree (RRT)  & RRT \cite{lavalle1998rapidly} and its derivatives are widely used sampling-based methods. These methods randomly sample in the configuration space or workspace to generate new tree vertices and connect the collision-free vertices as tree edges. In addition, these methods can consider the kinematic constraints (i.e., curvature limitations) during MP.  \\ 
b. & Probabilistic RoadMap* (PRM*)  & A probabilistic roadmap is a network graph of possible paths in a given map based on free and occupied spaces \cite{geraerts2004comparative, karaman2011sampling}. PRM* takes random samples from the robot's configuration space, tests them for whether they are in the free space, and uses a local planner to attempt to connect these configurations to other nearby configurations. Then, the starting and goal configurations are added in, and a graph search algorithm is applied to the resulting graph to determine a path between these two configurations.  \\
\midrule
3. & \textbf{Optimization-based} &  \\ \midrule
a. &  Mathematical Model  & \ac{MP} can be formulated as a path optimization problem with constraints on the robot model, such as its kinematic model \cite{raja2012optimal}. \\ 
b. & Evolutionary algorithms  & Evolutionary algorithms use bio-inspiration to find approximate solutions to difficult optimization problems. \cite{raja2012optimal}. \ac{ACO} is one of the population-based metaheuristic algorithms \cite{dorigo2006ant}. Artificial ants incrementally build solutions biased by a pheromone model, i.e.  a set of parameters associated with graph components (either nodes or edges) whose values are modified at runtime by the ants. \\ \midrule
4. &  \textbf{Learning-based} &  \\ \midrule
a. & Learning from Demonstrations (LfD) & LfD is the paradigm where an agent acquires new skills by learning to imitate an expert. LfD approach is compelling when ideal behavior cannot be easily scripted, nor defined easily as an optimization problem, but can be demonstrated \cite{ravichandar2020recent}. \\ 
b. & Reinforcement Learning (RL) & In RL, an agent learns to maximise a specific reward signal through trial and error interaction with the environment by taking actions and observing the reward \cite{sutton2018reinforcement}.\\
\bottomrule
\end{longtable}
\endgroup
\subsubsection{Node-based algorithms}

\begin{figure}[tpb]
    \centering
    \includegraphics[width=.7\linewidth]{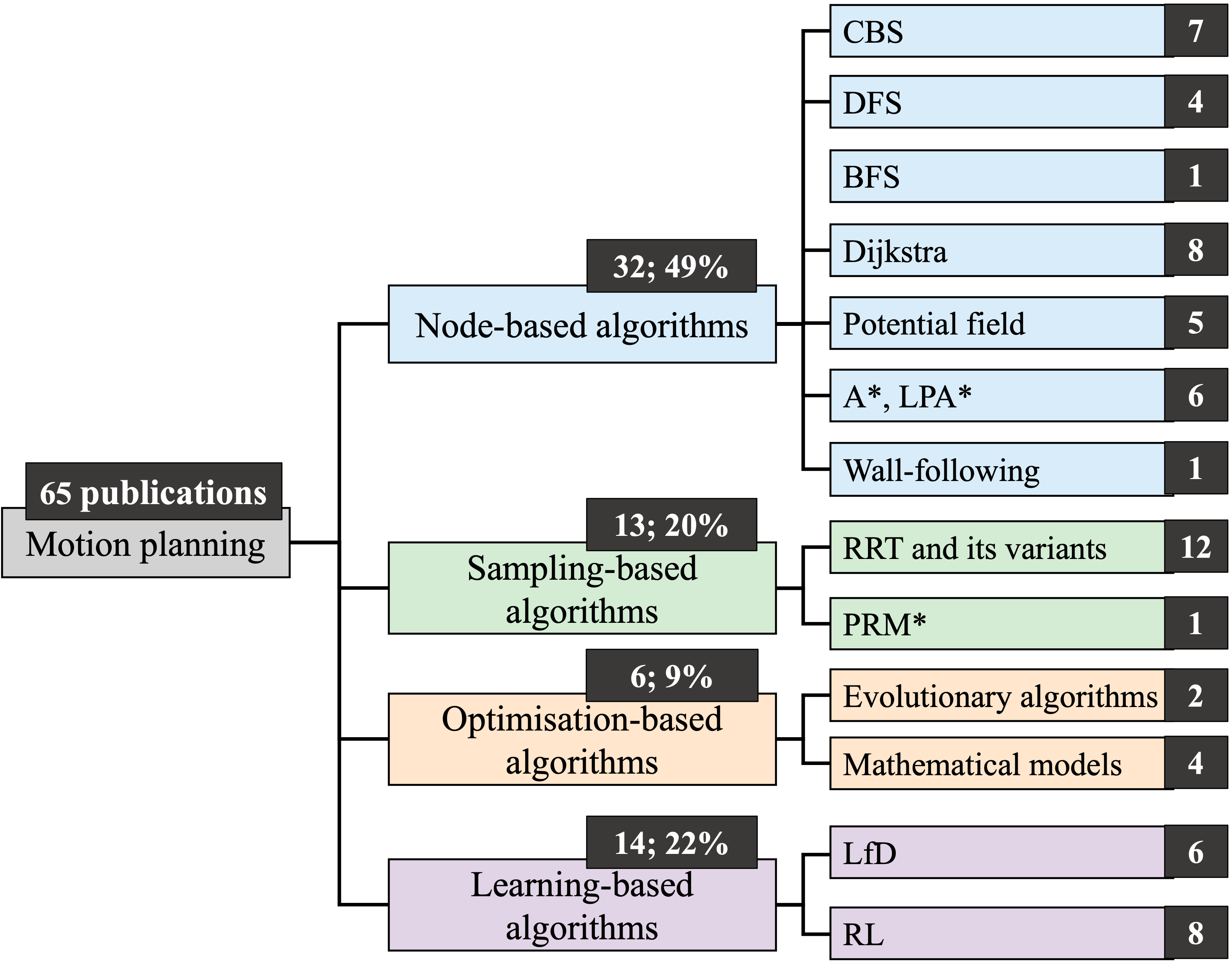}
    \caption{Classification of IPEI MP methods for continuum robots found in literature}
    \label{fig:taxonomy}
\end{figure}

Node-based algorithms use an information structure to represent the environment map and are commonly used for navigation assistance \cite{yang2016survey}. Table~\ref{tab:reflist} shows different MP works for \ac{IPEI} that exploit node-based methods. As schematised in Fig.~\ref{fig:taxonomy}, algorithms that have been adopted here are \ac{CBS}, \ac{DFS}, \ac{BFS}, Dijkstra, potential field, A*, \ac{LPA*}, and wall-following. 

\paragraph{Centerline-based Structure} 
Geiger \textit{et al.} extracts the 3D skeleton for bronchoscopy planning by computing the skeleton of the segmented structure and then converting this skeleton into a hierarchical tree model of connected branches \cite{Geiger_2005}.  
S{\' {a}}nchez \textit{et al.} \cite{S_nchez_2016} obtains the skeleton of the bronchial anatomy via the fast marching method firstly and then defines the skeleton branching points as a binary tree (B-tree). S{\' {a}}nchez' study gives a path corresponding to a sequence of nodes traversing the B-tree. Intraoperatively, a geometry likelihood map is used to match the current exploration to the path planned preoperatively.
The airway centerlines serve as the natural pathways for navigating through the airway tree. They are represented by a discrete set of airway branches in \cite{khare2015hands}. Starting with each target \ac{ROI} associated airway route, the method from Khare \textit{et al.}~\cite{khare2015hands} automatically derives a navigation plan that consists of natural bronchoscope manoeuvres abiding by the rotate-bend-advance paradigm learned by physicians during their training. This work is evaluated both in phantoms and in a human study. 

Wang \textit{et al.} developed a method to build a navigation information tree based on the vasculature's centerline for catheterisation \cite{Wang_2011}. The authors made a tree structure assuming the vascular system was rigid and interrogated the tree to find the nearest node during intraoperative navigation. The navigation experiments were carried out on a resin vessel phantom. Another study proposed a 3D vasculature's centerline extraction approach via a Voronoi diagram  \cite{yang2014centerlines}. It treated the centerlines as the minimal action paths on the Voronoi diagrams inside the vascular model surface. The experimental results show that the approach can extract the centerlines of the vessel model. 
Further Zheng \textit{et al.} \cite{zheng20183d} firstly proposed to extract the preoperative 3D skeleton via a parallel thinning algorithm for medical axis extraction \cite{kerschnitzki2013architecture}. Secondly, they proposed to use a graph matching method to establish the correspondence between the 3D preoperative and 2D intraoperative skeletons, extracted from 2D intraoperative fluoroscopic images. However, the proposed graph matching is sensitive to topology variance and transformation in the sagittal and transverse planes. A recent study on transnasal exploration proposed central path extraction algorithm based on pre-planning for the roaming area \cite{yudong2021rapid}.

Nevertheless, a common disadvantage of work available in the literature describing this approach is that they focus on constructing an information structure, but path exploration inside the information structure is not mentioned \cite{Geiger_2005, S_nchez_2016, khare2015hands, Wang_2011, yang2014centerlines, zheng20183d}. Specifically, the tree structure is built, but the path solution is not generated autonomously through a graph search strategy, especially when there are multiple path solutions simultaneously. 

\paragraph{Depth First Search} As an extended method to travel the tree formed in \cite{khare2015hands}, the studies by Zang \textit{et al.} implement a route search strategy of \ac{DFS} for an integrated endobronchial ultrasound bronchoscope, exploring a graph by expanding the most promising node along the depth \cite{zang2019optimal}, \cite{zang2021image}. In another study by Gibbs \textit{et al.}, a \ac{DFS} to view sites is regarded as the first phase search, followed by a second search focusing on a \ac{ROI} localisation phase and a final refinement to adjust the viewing directions of the bronchoscope \cite{gibbs2013optimal}. A \ac{DFS} approach is also developed in Huang \textit{et al.} for endovascular interventions \cite{huang2011interactive}. Instead of considering path length as node weights in the typical \ac{DFS} approach, this work defines the node weights as an experience value set by doctors.

The search time and the planned path are significantly dependent on the order of nodes in that same graph layer. Even though a \ac{DFS} approach can search for a feasible path by first exploring the graph along with the depth, it does not ensure that the first path found is the optimal path.

\paragraph{Breadth First Search}
The BFS algorithm was employed in \cite{fischer2022using} for a magnetically-actuated catheter to find a path reaching the target along vascular centerline points. However, the \ac{BFS} algorithm would take much more time to find a solution in a complex vascular environment with multi-branches.

\paragraph{Dijkstra} A graph structure based on vasculature's centerlines that are determined using a volume growing and a wavefront technique is designed by Schafer \textit{et al.} in \cite{Schafer_2007}. The optimal path is then determined using the shortest path algorithms from Dijkstra. However, Schafer \textit{et al.} assume that the centerline points are input as an ordered set, which would be a strict assumption. Moreover, they only report the scenario of a single lumen without branches, which does not reflect the advantages of the Dijkstra algorithm. A similar method but in a backward direction is presented by Egger \textit{et al.} \cite{egger2007}. This work determines an initial path by Dijkstra. Users define initial and destination points. After that, the initial path is aligned with the blood vessel, resulting in the vasculature's centerline. However, this methodology is not fully autonomous, and it involves manually tuned parameters. Another work extracts the centerline and places a series of guiding circular workspaces along the navigation path that are perpendicular to the path \cite{liu2010vitro}. The circular planes jointly form a safe cylindrical path from the start to the target. The Dijkstra algorithm is implemented to find the minimal cumulative cost set of voxels within the airway tree for bronchoscope navigation \cite{gibbs20073d,gibbs2008integrated} and find the shortest path along vasculature's centerlines \cite{qian2019towards}, \cite{schegg2022automated, cho2021image}.

Compared to \ac{DFS}, Dijkstra keeps tracking and checking the cost until it reaches the target. So there is a higher possibility of getting a better solution. Nevertheless, these researches still focus on tracking anatomical centerlines that are difficult to follow precisely and often not desirable. Because aligning the instrument tip with the centerline may call for excessive forces at more proximal points along the instrument's body where contact with the anatomy occurs.

\begin{figure}[tpb]
\centering  
\includegraphics[width=0.8\linewidth]{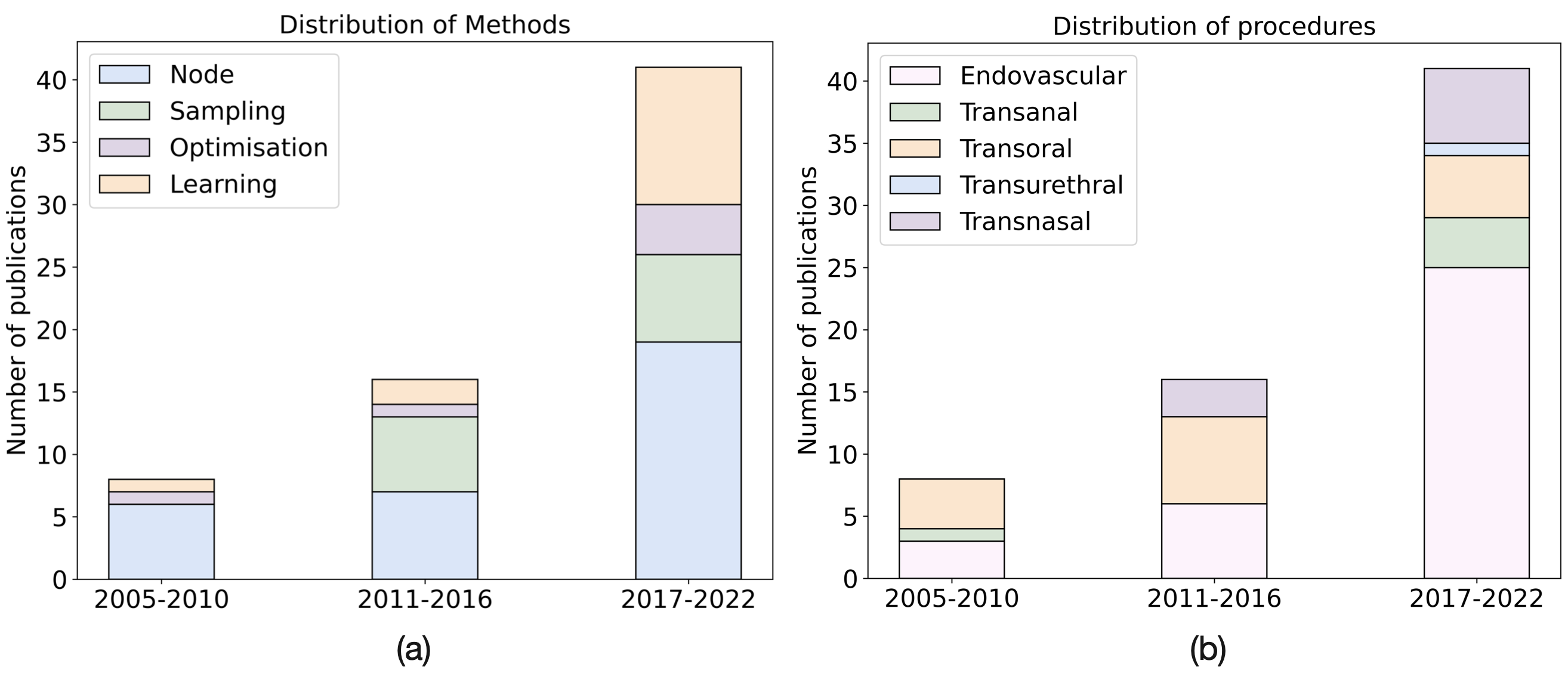}
\caption{Chronological development of endoluminal navigation. (a) MP approaches (b) The targeted \ac{IPEI} procedures. Until 2010, the majority of studies have implemented node-based and sampling algorithms for MP. While lately, with the exponential increase in computational resources, the field is transitioning towards learning-based methods.}
\label{fig:pub}
\end{figure}

\paragraph{Potential field} The work by Rosell \textit{et al.} \cite{rosell2012motion} computes the potential field over grids based on the L1 distance to obstacles. It is used to search a path by wavefront propagation for bronchoscopy. Rosell's approach considers the geometry and kinematic constraints while selecting the best motion according to a cost function. Yang \textit{et al.} \cite{yang2019path} extract centerlines via a distance field method, establish and navigate the tree after that.
However, the authors only considered the curvature constraint at 180$^{\circ}$ turns along vasculature's centerlines and assumed that all the path points have the same Y coordinate. Martin \textit{et al.} \cite{martin2020enabling} employ a potential field approach by defining an attractive force from the endoluminal image centre mass to the colon centre mass. 
A linear translation between the colon centre mass and the image centre is reconstructed and regarded as the linear motion of the colonoscope tip. This work is validated both in the synthetic colon and pig colon (\textit{in-vivo}). A similar approach is followed by Zhang \textit{et al.} where a robotic endoscope platform is employed to bring surgical instruments at the target site \cite{zhang2020enabling}. Girerd \textit{et al.}~\cite{girerd2020slam} use a 3D point cloud representation of a tubular structure and compute a repulsive force to ensure that the concentric tube needle tip remains inside the contour. 

The Potential field has an advantage in local planning by maintaining the centre of the image close to the centre of the cross-section of the lumen or the vessels. Nevertheless, it only considers a short-term benefit rather than global optimality during this local planning and might get stuck in a local minimum during global path planning.

\paragraph{A* and  Lifelong  Planning  A*} He \textit{et al.}~\cite{he2020} compute and optimise  endoscopic paths using the A* algorithm. The effectiveness of the preoperatively planned path is verified by an automatic virtual nasal endoscopy browsing experiment. Ciobirca \textit{et al.} search shortest airway paths through voxels of a bronchus model using the A* algorithm \cite{ciobirca2018new}. They claimed that this method could potentially improve the diagnostic success rate with a system for tracking the bronchoscope during a real procedure. However, this statement has not been validated yet. 
Some studies proposed a path planning method for \ac{CTRs} in brain surgery. The authors of these studies build a nearest-neighbour graph and use \ac{LPA*} algorithm for efficient replanning to optimise the insertion pose \cite{niyaz2018following, niyaz2019optimizing}. Compared to A*, \ac{LPA*}~\cite{koenig2004design} can reuse information from previous searches to accelerate future ones. Ravigopal \textit{et al.} proposed a modified hybrid A* search algorithm to navigate a tendon-actuated coaxially aligned steerable guidewire robot along a pre-computed path in 2D vasculature phantoms under C-arm fluoroscopic guidance \cite{ravigopal2021automated}. Recently, Huang \textit{et al.} showed colon navigation using a real-time heuristic searching method, called Learning real-time A* (LRTA*) \cite{huang2021autonomous}. 
LRTA* with designed directional heuristic evaluation shows efficient performance in colon exploration compared to BFS and DFS. Directional biasing avoids the need for unnecessary searches by constraining the next state based on local trends.

A* and LPA* use heuristic information to reach the goal. The first is commonly used for static environments, while the second can adapt to changes in the environment. They can converge very fast while ensuring optimality because both the cost from the start and the cost to the goal are taken into account. But their execution performance depends on the accuracy of the heuristic information. If inaccurate heuristic information is employed, searching in non-optimal directions severely affects its performance. 

\paragraph{Wall-following} The study in \cite{fagogenis2019autonomous} uses a wall-following algorithm to assist catheter navigation. Fagogenis \textit{et al.} \cite{fagogenis2019autonomous} employ haptic vision to accomplish wall-following inside the blood-filled heart for a catheter. The wall-following algorithm could be considered an efficient navigation approach if there are few feasible routes to reach the target state. Otherwise, the solution of a wall-following algorithm cannot ensure optimality.

\subsubsection{Sampling-based algorithms}~\\
\indent As observable in Table~\ref{tab:reflist}, different works, in the context of MP for \ac{IPEI}, exploit sampling-based methods. As schematised in Fig.~\ref{fig:taxonomy}, algorithms based on \ac{RRT} and its variants and \ac{PRM*} have been proposed.

\paragraph{Rapidly-exploring Random Tree and its variants} Some studies compare several \ac{RRT}-based algorithms looking for the optimal option for the virtual bronchoscopy simulator, such as \ac{RRT}, \ac{RRT}-connect, dynamic-domain \ac{RRT}, and \ac{RRT}-Connect with dynamic-domain \cite{aguilar2017rrt, aguilar2017virtual}. Results reveal that the RRT-Connect with Dynamic Domain is the optimal method requiring the minimum number of samples and computational time for finding the solution path. Fellmann \textit{et al.} use a collision-free path via \ac{RRT} as a baseline \cite{fellmann2015implications}. Then different trajectory generation strategies are applied and evaluated. Inside the narrow and straight nasal passage, Fellmann \textit{et al.} report that the best strategy is synchronous point-to-point. However, that strategy could become infeasible as the distance between intermediate configurations increases. Kuntz \textit{et al.} \cite{kuntz2015motion} introduce a \ac{RRT}-based algorithm in a three-step planning approach for a novel transoral lung system consisting of a bronchoscope, a CTR, and a bevel-tip needle. Their approach considers the ability of needle steering during path planning. Kuntz \textit{et al.} demonstrate the motion planner's ability to respect a maximum needle steering curvature. The time to find a motion plan significantly depends on the steering capability and the target location.

The study in \cite{guo2021training} implements an improved \ac{RRT} algorithm for cerebrovascular intervention. The expansion direction of the random tree is a trade-off between the new randomly sampled node and the target. This strategy can improve the convergence speed of the algorithm, even if catheter constraints are not considered.  

Alterovitz \textit{et al.} \cite{alterovitz2011rapidly} proposed a \ac{RRM} method that initially explores the configuration space like \ac{RRT}. Once a path is found, \ac{RRM} uses a user-specified parameter to weigh whether to explore further or to refine the explored space by adding edges to the current roadmap to find higher-quality paths in the explored space. Their method is presented for \ac{CTRs} in a tubular environment with protrusions as bronchus. Some studies develop the RRM method and improve it with more accurate mechanics-based models in a skull base surgery scenario and static lung bronchial tubes for \ac{CTRs} respectively \cite{torres2011motion}, \cite{torres2012task}. In Torres \textit{et al.} \cite{torres2011motion}, the planner required \SI{1077}{\s} to get a motion plan that avoids bone, critical blood vessels and healthy brain tissue on the way to the skull base tumour. The same authors extend the previous studies in \cite{torres2014interactive} by proposing a modified \ac{RRG} method that computes motion plans at interactive rates. This work improves the computation cost and allows replanning when the robot tip position changes. However, generating such a roadmap requires an extensive amount of computation. Therefore, the method could behave well in a static environment but not in deformable lumens.

Fauser \textit{et al.} use the formulation of RRT-connect (or bi-directional RRT, Bi-RRT) introduced earlier by them \cite{fauser2018planning} to solve a common MP problem for instruments that follow curvature constrained trajectories \cite{fauser2018generalized}. In \cite{fauser2019optimizing}, Fauser \textit{et al.} implement the RRT-connect algorithm for a catheter in a 3D static aorta model, under the allowed maximal curvature \SI{0.1}{\per\mm}. Further extension of this work proposes path replanning from different robot position states along the initial path starting from the descending aorta to the goal in the left ventricle \cite{fauser2019planning}.

\paragraph{Probabilistic RoadMap*} Kuntz \textit{et al.} propose a method based on a combination of a \ac{PRM*} method and local optimisation to plan motions in a point cloud representation of a nasal cavity anatomy \cite{kuntz2019planning}. The limitation is that the anatomy model is only updated within the visible region of the endoscope, while deformations of the rest of the anatomy are not considered. If tissue deformation is negligible, this planning method could be used for intraoperative planning. Otherwise, the deformations of the overall model must be considered beforehand.

\subsubsection{Optimisation-based algorithm}~\\
\indent \ac{MP} can be formulated as an optimisation problem and solved by numerical solvers \cite{raja2012optimal}. Moreover, these methods can be programmed to consider also the robotic kinematics.

\paragraph{Mathematical model} An optimisation-based planning algorithm that optimises the insertion length and orientation angle of each tube for a CTR with five tubes is proposed by Lyons \textit{et al.} \cite{lyons2010planning}. Firstly, the authors formulate the MP problem as a non-linear constrained optimisation problem. Secondly, the constraint is moved to the objective function, and the problem is converted to a series of unconstrained optimisation problems. Lastly, the optimal solution is found using the \ac{L-BFGS} algorithm \cite{liu1989limited} and Armijo's Rule \cite{bazaraa2013nonlinear}. The robot kinematics is modelled using a physically-based simulation that incorporates beam mechanics. This work is evaluated in simulation on a patient's lung anatomy. However, the computational time of the proposed method is high, 

\begin{landscape}
\begin{footnotesize}
\begin{longtable}[c]{p{.14\linewidth}p{.11\linewidth}p{.13\linewidth}p{.14\linewidth}p{.05\linewidth}p{.19\linewidth}p{.08\linewidth}}
\caption{Summary of motion planning methods for IPEI navigation }\label{tab:reflist}\\
\hline
\rowcolor{lightgray}\textbf{Ref.} & \textbf{Procedure} & \textbf{Method} & \textbf{Robot} & \textbf{Kinematics}  & \textbf{Validation} &  \textbf{Environment} \\
\hline

\endfirsthead
\multicolumn{7}{l}%
{\tablename\ \thetable\ -- \textit{Continued from previous page}} \\
\hline
\rowcolor{lightgray}\textbf{Ref.} & \textbf{Procedure} & \textbf{Method} & \textbf{Robot} & \textbf{Kinematics}  & \textbf{Validation} &  \textbf{Environment} \\
\hline
\endhead
\hline \multicolumn{7}{r}{\textit{Continued on next page}} \\
\endfoot
\hline
\endlastfoot

\hline
\rowcolor{lightyellow}\multicolumn{7}{l}{\textbf{Level 1: Navigation assistant}} \\
\rowcolor{lightblue}\multicolumn{7}{l}{\textit{\textbf{Node-based:}}} \\
\cite{Geiger_2005} Geiger 2005 & Transoral & CBS & Bronchoscope & N & in-silico (3D pulmonary vessels) & Static \\
\cite{Schafer_2007} Schafer 2007 & Endovascular & Dijkstra & Guidewire & N & in-vitro (3D cardiovascular) & Static \\
\cite{egger2007} Egger 2007 & Endovascular & Dijkstra & Catheter & N & in-silico (3D aorta) & Static \\
\cite{gibbs20073d} Gibbs 2007 & Transoral & Dijkstra & Bronchoscope & N & in-silico (3D bronchus) & Static \\
\cite{gibbs2008integrated} Gibbs 2008 & Transoral & Dijkstra & Bronchoscope & N & in-silico (3D bronchus) & Static \\
\cite{liu2010vitro} Liu 2010 & Endovascular & Dijkstra & Catheter & N & in-vitro (3D aorta) & Static \\
\cite{Wang_2011} Wang 2011 & Endovascular & CBS & Catheter & N & in-vitro (3D resin vessel) & Static \\
\cite{huang2011interactive} Huang 2011 & Endovascular & DFS & Guidewire & N & in-silico (3D aorta) & Static \\
\cite{rosell2012motion} Rosell 2012 & Transoral & Potential field & Bronchoscope & Y & in-silico (3D tracheobronchial) & Static \\
\cite{gibbs2013optimal} Gibbs 2013 & Transoral & DFS & Bronchoscope & Y & in-vivo (3D bronchus) & Dynamic \\
\cite{yang2014centerlines} Yang 2014 & Endovascular & CBS & Guidewire & N & in-silico (3D aorta) & Static \\
\cite{khare2015hands} Khare 2015 & Transoral & CBS & Bronchoscope & Y & in-vivo (3D bronchus) & Dynamic \\
\cite{S_nchez_2016} S{\'{a}}nchez 2016 & Transoral & CBS & Bronchoscope & N & in-silico (3D bronchus) & Static \\
\cite{zheng20183d} Zheng 2018 & Endovascular & CBS & Catheter & N & in-vitro (3D aorta) & Dynamic \\
\cite{ciobirca2018new} Ciobirca 2018 & Transoral & A* & Bronchoscope & N & in-silico (3D bronchus) & Static\\
\cite{niyaz2018following} Niyaz 2018 & Transnasal & LPA* & Concentric tube robot & Y & in-silico (3D nasal cavity) & Static \\
\cite{niyaz2019optimizing} Niyaz 2019 & Transnasal & LPA* & Concentric tube robot & Y & in-silico (3D nasal cavity) & Static \\
\cite{zang2019optimal} Zang 2019 & Transoral & DFS & Bronchoscope & Y & in-vivo (3D bronchus) & Dynamic \\
 \cite{yang2019path} Yang 2019 & Transurethral & Potential field & Ureteroscope & Y & in-silico (3D ureter) & Static \\
 \cite{fagogenis2019autonomous} Fagogenis 2019 & Endovascular & wall-following & Concentric tube robot & Y & in-vivo (3D cardiovascular) & Dynamic \\
\cite{he2020} He 2020 & Transnasal & A* & Endoscope & N & in-silico (3D nasal cavity) & Static \\
\cite{zang2021image} Zang 2021 & Transoral & DFS & Bronchoscope & Y & in-vivo (3D bronchus) & Static \\
\cite{huang2021autonomous} Huang 2021 & Transanal & LRPA* & Colonoscope & Y & in-vitro (2D colon) & Dynamic \\
\cite{yudong2021rapid} Wang 2021 & Transnasal & CBS & Endoscope & N & in-silico (3D nasal cavity) & Static \\
\cite{ravigopal2021automated} Ravigopal 2021 & Endovascular & Hybrid A* & Robotic guidewire & Y & in-vitro (2D vessel) & Static \\
\hline
\rowcolor{lightgreen}\multicolumn{7}{l}{\textit{\textbf{sampling-based:}}}\\
\cite{alterovitz2011rapidly} Alterovitz 2011 & Transoral & RRM & Concentric tube robot & Y & in-silico (3D bronchus) & Static \\
\cite{torres2011motion} Torres 2011 & Transnasal & RRM & Concentric tube robot & Y & in-silico (3D nasal cavity) & Static \\
\cite{torres2012task} Torres 2012 & Transoral & RRM & Concentric tube robot & Y & in-silico (3D bronchus) & Static \\
\cite{torres2014interactive} Torres 2014 & Transnasal & RRG & Concentric tube robot & Y & in-silico (3D nasal cavity) & Static \\
\cite{fellmann2015implications} Fellmann 2015 & Transnasal & RRT & Concentric tube robot & Y & in-silico (3D nasal cavity) & Static \\
\cite{kuntz2015motion} Kuntz 2015 & Transoral & RRT & Steerable needle & Y & in-silico (3D bronchus) & Static \\
\cite{aguilar2017rrt} Aguilar 2017 & Transoral & bi-RRT & Bronchoscope & Y & in-silico (3D bronchus) & Static \\
\cite{aguilar2017virtual} Aguilar 2017 & Transoral & bi-RRT & Bronchoscope & Y & in-silico (3D bronchus) & Static \\
\cite{fauser2018generalized} Fauser 2018 & Endovascular & bi-RRT & Catheter & Y & in-silico (3D vena cava) & Static \\
\cite{fauser2019optimizing} Fauser 2019a & Endovascular & bi-RRT & Steerable guidewire & Y & in-silico (3D aorta) & Static \\
\cite{fauser2019planning} Fauser 2019b & Endovascular & bi-RRT & Steerable guidewire & Y & in-silico (3D aorta) & Static \\
\cite{kuntz2019planning} Kuntz 2019 & Transnasal & PRM* & Concentric tube robot & Y & in-silico (3D nasal cavity) & Dynamic \\
\cite{guo2021training} Guo 2021 & Endovascular & RRT & Catheter & N & in-silico (cerebrovascular) & Static \\
\hline
\rowcolor{lightorange}\multicolumn{7}{l}{\textit{\textbf{Optimisation-based:}}}\\
\cite{lyons2010planning} Lyons 2010 & Transoral endotracheal & Mathematical model & Concentric tube robot & Y & in-silico (3D bronchus) & Static \\
\cite{gao2015three} Gao 2015 & Endovascular & ACO & Catheter & Y & in-silico (3D lower limb arteries) & Static \\
\cite{qi2019kinematic} Qi 2019 & Endovascular & Mathematical model & Continuum robot & Y & in-vitro (blood vessels) & Static \\
\cite{li2021path} Li 2021 & Endovascular & GA & Catheter & Y & in-silico (3D aorta and coronaries) & Static \\
\cite{guo2021design} Guo 2021 & Endovascular & Mathematical model & Catheter & Y & in-silico, in-vitro (3D vessel model) & Static \\
\cite{abah2021image} Abah 2021 & Endovascular & Mathematical model & Catheter & Y & in-vitro (3D cerebrovascular) & Static \\
\hline
\rowcolor{lightviolet}\multicolumn{7}{l}{\textit{\textbf{Learning-based:}}}\\

\cite{zhao2022surgical} Zhao 2022 & Endovascular & LfD using GAN & Guidewire & N & in-vitro (3D vessel model) & Static \\
\cite{meng2021evaluation} Meng 2021 & Endovascular & RL & Catheter & N & in-silico (3D aorta) & Static \\
\hline

\rowcolor{lightyellow}\multicolumn{7}{l}{\textbf{Level 2: Navigation using waypoints}} \\
\rowcolor{lightviolet}\multicolumn{7}{l}{\textit{\textbf{Learning-based:}}}\\
\cite{trovato2010development} Trovato 2010 & Transanal & RL & Fibre optic endoscope & N & ex-vivo (3D swine colon) & Dynamic \\
\cite{Rafii_Tari_2013} Rafii-Tari 2013 & Endovascular & LfD using GMM & Catheter & N & in-vitro (3D aorta) & Static \\
\cite{rafii2014hierarchical} Rafii-Tari 2014 & Endovascular & LfD using H+HMM & Catheter & Y & in-vitro (3D aorta) & Static \\
\cite{Chi_2018} Chi 2018a & Endovascular & LfD using DMPs & Catheter & Y & in-vitro (3D aorta) & Dynamic \\
\cite{Chi_2018_2} Chi 2018b & Endovascular & LfD using GMMs & Catheter & N & in-vitro (3D aorta) & Static \\
\cite{chi_2020} Chi 2020 & Endovascular & LfD using GAIL & Catheter & N & in-vitro (3D aorta) & Static \\
\hline
\rowcolor{lightyellow}\multicolumn{7}{l}{\textbf{Level 3: Semi-autonomous navigation}} \\
\rowcolor{lightblue}\multicolumn{7}{l}{\textit{\textbf{Node-based:}}} \\
\cite{qian2019towards} Qian 2019 & Endovascular & Dijkstra & Guidewire & N & in-vitro (3D femoral arteries, aorta) & Static \\
\cite{girerd2020slam} Girerd 2020 & Transnasal & Potential field & Concentric tube robot & Y & in-silico (3D nasal cavity), in-vitro (origami tunnel) & Static \\
\cite{martin2020enabling} Martin 2020 & Transanal & Potential field & Endoscope & N & in-vivo (3D colon) & Dynamic \\
\cite{zhang2020enabling} Zhang 2020 & Transanal & Potential field & Endoscope & Y & in-vitro (2D colon model) & Dynamic \\
\cite{cho2021image} Cho 2021 & Endovascular & Dijkstra & Guidewire & N & in-vitro (2D vessel) & Static \\
\cite{fischer2022using} Fischer 2022 & Endovascular & BFS & Catheter & N & in-vitro (2D vessel) & Static \\
\cite{schegg2022automated} Schegg 2022 & Endovascular & Dijkstra & Guidewire & N & in-silico (3D coronary arteries) & Static \\
\hline
\rowcolor{lightviolet}\multicolumn{7}{l}{\textit{\textbf{Learning-based:}}}\\
\cite{you2019automatic} You 2019 & Endovascular & RL & Catheter & N & in-vitro (3D heart) & Static \\
\cite{behr2019deep} Behr 2019 & Endovascular & RL & Catheter & N & in-vitro (2D vessel) & Static \\
\cite{karstensen2020autonomous} Karstensen 2020 & Endovascular & RL & Catheter & N & in-vitro (2D vessel) & Static \\ 
\cite{kweon2021deep} Kweon 2021 & Endovascular & RL & Guidewire & N & in-vitro (2D coronary artery) & Static \\
\cite{pore2022colonoscopy} Pore 2022 & Transanal & RL & Endoscope & N & in-silico (3D colon) & Dynamic \\
\cite{karstensen2022learning} Karstensen 2022 & Endovascular & RL & Guidewire & N & ex-vivo (2D venous system) & Dynamic \\

\hline
\end{longtable}
\end{footnotesize}
\end{landscape}

which restricts the possibility of applying this method to real-time scenarios. Moreover, the authors manually define the skeleton and treat the structure as a rigid body, confining its applicability.

An inverse kinematics MP method for continuum robots is expressed as an optimisation problem based on the backbone curve method by Qi \textit{et al.} \cite{qi2019kinematic}. The technique minimises the distance to the vasculature’s centerline under kinematic constraints independently during each step without considering a long-term cumulative cost. Therefore, optimal inverse kinematics that does not consider the past and future phases might not be globally optimal.

Guo \textit{et al.}~\cite{guo2021design} employed directional modeling of a teleoperated catheter and proposed a hybrid evaluation function to find the optimal trajectory. This work conducted wall-hit experiments and compared the response time of obstacle avoidance with and without path planning. However, the optimal solution is obtained with an exhaustive enumeration, which is a computationally expensive solution. Abah \textit{et al.}~\cite{abah2021image} consider the path planning as a nonlinear least-squares problem to minimize the passive deflection of the steerable catheter. It is achieved by matching the shape of the steerable segment as closely as possible to the centerline of the cerebrovascular. Nevertheless, the centerline might not be the optimal reference route for steerable catheters.

\paragraph{Evolutionary algorithms} An improved \ac{ACO} method is proposed to plan an optimal vascular path with overall consideration of factors such as catheter diameter, vascular length, diameter, as well as curvature and torsion \cite{gao2015three}. The associated computational time varied from \SI{2}{\s} to \SI{30}{\s}, with an average value \SI{12.32}{\s}. The high computational time cost limits its application in real-time scenarios.
Li \textit{et al.}~\cite{li2021path} proposed a fast path planning approach under the steerable catheter curvature constraint via a local \ac{GA} optimisation. The reported results showed the planner’s ability to satisfy the robot curvature constraint while keeping a low computational time cost of $0.191\pm0.102$s.

\subsubsection{Learning-based algorithms}~\\
\indent Learning-based methods are a viable candidate for real-time MP. These methods use statistical tools such as Artificial Neural Networks, Hidden Markov models (HMMs), and dynamical models to map perceptual and behaviour spaces. In the context of this article, we identified Learning from Demonstrations (LfD) and \ac{RL} approaches as sub-fields of learning methods.

\paragraph{Learning from Demonstrations}
Rafii-Tari \textit{et al.} provides a system for human-robot collaboration for catheterisation \cite{rafii2014hierarchical}. The catheterisation procedure is decomposed manually into a series of catheter movement primitives. These primitive motions are modelled as \ac{HMMs} and are learnt using a \ac{LfD} approach. Additionally, a high-level HMM is learnt to sequence the motion primitives. Another system, proposed by the same authors, provides a semi-automated approach for navigation, in which guidewire manipulation is controlled manually, and catheter motion is automated by the robot \cite{Rafii_Tari_2013}. Catheter motion is modelled here using a \ac{GMM} to create a representation of temporally aligned phase data generated from demonstrations. Chi \textit{et al.} extend this work by showing subject-specific variability among type I aortic arches through incorporating the anatomical information obtained from preoperative image data \cite{Chi_2018_2}. In all the above methods, expectation maximisation was used to perform maximum-likelihood estimation to learn the model parameters.
Another study presents a \ac{LfD} method based on \ac{DMPs} \cite{Chi_2018}.
\ac{DMPs} are compact representations for motion primitives formed by a set of dynamic system equations \cite{saveriano2021dynamic}. The study uses \ac{DMPs} to avoid unwanted contact between the catheter tip and the vessel wall. \ac{DMPs} were trained from human demonstrations and used to generate motion trajectories for the proposed robotic catheterisation platform. 
The proposed methods can adapt to different flow simulations, vascular models, and catheterisation tasks. 
In a recent continuation of their prior study, Chi \textit{et al.} improves the \ac{RL} part by including model-free \ac{GAIL} loss that learns from multiple demonstrations of an expert \cite{chi_2020}. In this work, the catheterisation policies adapt to the real-world setup and successfully imitate the task despite unknown simulated parameters such as blood flow and tissue-tool interaction. Zhao \textit{et al.} proposed a \ac{GAN} framework by combining \ac{CNN} and \ac{LSTM} \cite{zhao2022surgical} to estimate suitable manipulation actions for catheterization. The \ac{DNN} is trained using experts' demonstration data and  evaluated in a phantom with a grey-scale camera simulating X-ray imaging.

\paragraph{Reinforcement Learning}
Trovato \textit{et al.} developed a hardware system for a robot colonic endoscope. It showed that the voltage for propulsion could be controlled through classic \ac{RL} algorithms such as State-Action-Reward-State-Action (SARSA) and Q-learning that could determine the forward and backward motion \cite{trovato2010development}.
Existing state-of-the-art \ac{RL} algorithms use \ac{DNN} to learn from high-dimensional and unstructured state inputs with minimal feature engineering to accomplish tasks, called \ac{DRL} \cite{mnih2015human}.
Recently, Behr \textit{et al.} \cite{behr2019deep}, Karstensen \textit{et al.} \cite{karstensen2020autonomous} and Meng \textit{et al.} \cite{meng2021evaluation} proposed a closed-loop control system based on \ac{DRL}, which uses the kinematic coordinates of the guidewire tip and manipulator as input and outputs continuous actions for each degree of freedom for rotation and translation. \cite{karstensen2022learning} showed the translation in ex-vivo veins of a porcine liver.
To improve the previously closed-loop control, You \textit{et al.} \cite{you2019automatic} and Kweon \textit{et al.} \cite{kweon2021deep} automate control of the catheter using \ac{DRL} based on image inputs in addition to the kinematic information of the catheter. The authors train a policy in a simulator and show its translation to a real robotic system. The real robotic experiments are carried out using the tip position from an electromagnetic sensor sent to the simulator to realise the virtual image input.

For transanal \ac{IP}, Pore \textit{et al.} proposed a deep visuomotor control to map the endoscopic images to the control signal \cite{pore2022colonoscopy}. The study reported efficient colon navigation in various in-silico colon models and better navigation performance compared to experts in terms of overall trajectory properties.
Other efforts where some applications of \ac{DRL} are emerging is tracheotomy. For example, Athiniotis \textit{et al.} uses a snake-like clinical robot to navigate down the airway \cite{athiniotisdeep} autonomously. In this work, they employ a \ac{DQN} based navigation policy that utilises images from a monocular camera mounted on its tip. The system serves as an assistive device for medical personnel to perform endoscopic intubation with minimal human intervention.

\subsection{Limitations of present MP methods} \label{sec:limitations}
MP is a key ingredient in enabling autonomous navigation. However, it suffers some limitations that hinder their universal application in IPEI procedures. 
In this section, we identify the limitations of the aforementioned MP methods.

\textit{Node-based}: The searching strategy of node-based algorithms is based on specific cost functions. The optimality and
\noindent completeness of the solution obtained using this strategy could be guaranteed. However, (i) node-based algorithms usually lack the consideration to satisfy robot capability during MP, such as robots' kinematic constraints; (ii) the uncertainty of sensing is rarely considered; (iii) the proposed methods are only applied in rigid environments, tissue deformations during procedures are not incorporated; (iv) node-based algorithms usually rely on the thorough anatomical graph structures. Accurate reconstructions of the anatomical environment in the preoperative phase are needed to build the data structure and  search inside it. The mentioned limitations reduce the usability of these methods. In theory, they may work, but in
practice, they are difficult to be applied for autonomous real-time navigation in real-life conditions.
\noindent 

\textit{Sampling and Optimisation based}: Sampling and optimisation-based approaches can account for the robot-specific characteristics. Nevertheless, the performance of these methods is affected significantly by the robot model. Moreover, especially for continuum soft robots, \cite{da2020challenges}, the modelling methods and soft constraints of obstacle collision are challenging and still under investigation. 
Sampling-based approaches reduce computational time compared to optimisation approaches but do not ensure the solution's optimality. The ``probabilistic" completeness of sampling-based methods is their intrinsic property due to their random sampling. In other words, finding a feasible path solution is not always guaranteed. 
Existing optimisation-based methods are time-consuming and mainly applied in static environments for preoperative MP. 
Hybrid methods that fuse multiple approaches could maximise their respective advantages.

\textit{Learning-based}: Learning-based methods implemented in robotics have been rising. However, current challenges associated with learning-based methods limit their universal application in the clinical scenario \cite{ibarz2021train}: 
One of the major concerns is safety \cite{garcia2015comprehensive}. Recently developed learning methods make use of \ac{DNN} that can show unpredictable behaviour for unseen data outside the training regime. Hence ensuring that the \ac{DNN} never makes decisions that can cause a safety violation is crucial \cite{pore2021safe, corsi2023constrained}. In addition, \ac{DNN}-based learning methods require a huge amount of training data due to their inherent complexity, the large number of parameters involved and the learning optimisation \cite{lecun2015deep}. Therefore, a massive amount of data need to be acquired, moved, stored, annotated and queried in an efficient way \cite{birkhoff2021review}. In the surgical domain, high-quality diverse information is rarely available \cite{kennedy2020computer}. Various groups have proposed shared standards for device integration, data acquisition systems and scalable infrastructure for data transmission, such as the CONDOR (Connected Optimized Network and Data in Operating Rooms) project (https://condor-h2020.eu/) and OR black box \cite{goldenberg2017using}. A general trend to overcome data limitations is through the use of simulators. However, it is challenging to generalise the knowledge gained through training in a simulator to a real situation, called the ``sim-to-real" reality gap. Discrepancies between reality and virtual environment occur due to modelling errors \cite{rusu2016sim}.
Notably, model-free \ac{DRL} is a widely popular way of learning goal-directed behaviours and has shown promising success in controlled robotic environments \cite{ibarz2021train}. Some commonly used algorithms include PPO (on-policy) \cite{schulman2017proximal}, SAC (off-policy) \cite{haarnoja2018soft}.
However, model-free \ac{DRL} suffers from several limitations. 
First, there is a need to design a reward function implicitly.
This need requires the developer to have domain knowledge of the dynamics of the environment \cite{ibarz2021train}, which is highly complex for deformable objects and tissues \cite{lin2020softgym, li2022position}.
Second, sensitivity to hyperparameters and under-optimised parameters can cause a significant difference in performance. Hence, a considerable amount of time has to be invested in tuning hyperparameters.
Third, learning from high-dimensional inputs such as images is challenging compared to low-dimensional state features such as robot kinematic data and has shown generalisation problems due to the high capacity of \ac{DNN} \cite{ibarz2021train}. 
Fourth, continuum robots such as endoscopes add to the dimensionality of the action space since they have a high number of \ac{DoFs} with complex architectures, compared to industrial robots \cite{dupont2022continuum}.
Some algorithm difficulty involves restricted policy search.

LfD is a preferred way to learn human gestures in the context of imitation learning \cite{pore2021learning}. However, a significant drawback of LfD methods is that they require many demonstrations to be adequately trained, which is unfeasible in clinical settings considering the time, resources and ethical constraints. Furthermore, LfD typically only enables the robot to become as good as the human's demonstrations since a large deviation of the policy from the demonstrated data could lead to unstable policy learning \cite{ghasemipour2020divergence}. 
\section{FUTURE DIRECTIONS}\label{sec:fd}

Navigation is one of the crucial interventional phases of an IPEI procedure.
The need for automation in IPEI navigation will increasingly support the adoption of novel MP techniques capable of working in unstructured and dynamic luminal environments. In this section, we describe the improvements in MP algorithms that have been applied in other robotics domains and can be extended to IPEI. Moreover, robot navigation relies on robot design and its sensing capabilities. Therefore, we discuss the essential robotics capabilities still missing to enable navigation systems with a higher level of autonomy (e.g., LoA 4).
\vspace{-3mm}
\subsection{Improvements in motion planning algorithms}

MP for continuum robots is a complex problem because many configurations exist with multiple internal \ac{DoFs} that have to be coordinated to achieve purposeful motion \cite{dupont2022continuum, burgner2015continuum}. 32 of 65 publications consider MP for the robot without considering its kinematics, as shown in Table~\ref{tab:reflist}.
Future studies need to focus on the robotic constraints for active MP. Moreover, replanning is required to adapt the current plan to deformable environments using sensorial information. The objective of replanning is to reduce the navigation error measured according to defined metrics. 
Therefore, the computational efficiency of MP becomes essential for real-time scenarios. This section highlights insights that can improve existing MP techniques, as discussed in Sec.~\ref{sec:mp}.

Some novel studies on the path planning of a steerable needle for neurosurgery could give some inspiration for \ac{IPEI}, as these studies considered curvature constraints of a robotic needle. Parallel path exploration is used in the Adaptive Fractal Trees (AFT) proposed for a programmable bevel-tip steerable needle \cite{liu2016fast}. This method uses fractal theory and Graphics Processing Units (GPUs) architecture to parallelize the planning process, and enhance the computation performance and online replanning, as demonstrated with simulated 3D liver needle insertions. An Adaptive Hermite Fractal Tree (AHFT) is later proposed, where the AFT is combined with optimised geometric Hermite curves that allow performing a path planning strategy satisfying the heading and targeting curvature constraints \cite{pinzi2019adaptive}. Although developed and tested only for a preoperative neurosurgical scenario, AHFT is well-suited for GPU parallelisation for rapid replanning.

Hybrid approaches can take advantage of individual methods to show enhanced performance and overcome the limitation of each method. The emerging learning-based approaches can be combined with other methods to overcome their limitations. For example, Wang \textit{et al.} propose a hybrid approach combining RL and RRT algorithms for MP in narrow passages \cite{wang2018learning}. Their method can enhance the local space exploration ability and guarantee the efficiency of global path planning. Some other authors also present hybrid MP methods for \ac{IPEI} navigation. For example, Meng \textit{et al.} propose a hybrid method using Breadth-First Search (BFS) and \ac{GA} for micro-robot navigation in blood vessels of rat liver, aiming to minimise the energy consumption \cite{meng2019motion}.

Optimisation-based methods are also an active area of research for obtaining an optimal preoperative plan under complex constraints. Particle Swarm Optimisation (PSO) is implemented by Granna \textit{et al.} for a concentric tube robotic system in neurosurgery \cite{granna2019computer}. Dynamic programming is employed for micro-robot path planning in rigid arteries under a minimum effort criterion \cite{pourmanda2019navigation}. However, the search space reduction technique for the constrained optimisation problem is essential for intraoperative MP. Howell \textit{et al.} propose an augmented Lagrangian trajectory optimiser solver for constrained trajectory optimisation problems in \cite{howell2019altro}. It handles general nonlinear state and input constraints and offers fast convergence and numerical robustness. For an \ac{IPEI} motion planner, an efficient optimisation solver with reduced search space would be potentially applied for intraoperative planning.

As demonstrated in Fig.~\ref{fig:pub}, the recent shift towards learning-based approaches has shown promising success. The guarantee of a provable behaviour using \ac{DNN} is still an open problem, and it is crucial to incorporate safety constraints for the automation of \ac{IPEI} navigation tasks to avoid hazardous actions. Some studies have proposed safe RL frameworks for safety-critical paradigms using barrier functions to restrict the robot actuation in a safe workspace \cite{cheng2019end, garcia2015comprehensive} and its behaviour is formally verified to guarantee safety \cite{pore2021safe, corsi2023constrained}.
Robot unsafe behaviour can also be generated due to large policy updates of gradient-based optimisation. Such large deviations can be limited by restricting the policy update in a trust region, leading to monotonic improvement in policy performance. Some works use f-divergences methods such as KL-divergence to constrain the policy search from being greedy \cite{schulman2017proximal}. To tackle the problem of high cost and danger of interacting with the environment and data inefficiency of existing \ac{DRL} methods, recent studies have explored offline RL that learns exclusively from static datasets of previously collected experiences \cite{prudencio2022survey}.

Commonly used model-free RL techniques do not consider the dynamics of the environment \cite{you2017model}. However, various complexities, such as pulsatile flow within the vasculature or nonlinear behaviour of the instrument, hinder the implementation of model-free algorithms and compel to simplify the problem sets. Thus, the future trend could involve implementing model-based approaches in endoluminal or endovascular environments \cite{thuruthel2018model}.  
Model-based approaches are sample-efficient and require less data for training \cite{liu2020efficient}. Hierarchical RL is another untapped field for long navigation tasks, which is oriented to subdivide the interventional phase into steps and applying specific policies to each. This approach better adapts to the specifications of each phase.
For example, in the case of \ac{IPEI} navigation, the complete navigation task could be subdivided and learnt incrementally \cite{pore2020simple}. Recently, curriculum learning has been proposed to learn in increasingly complex environments \cite{bengio2009curriculum}.

\vspace{-4mm}
\subsection{Robotic capabilities} \label{subsec:rob_cap}
Reaching higher LoA in navigation requires accurate control and enhanced shape-sensing capabilities. In this subsection, we discuss various missing capabilities in current IPEI robotic systems that hinder the development of a LoA~4 navigation system.

\subsubsection{Robotics actuation}
Continuum robots employed in \ac{IPEI} procedures are developed based on different designs and technologies. For instance, several continuum instruments use concentric tube mechanisms or multi-link systems \cite{burgner2015continuum, omisore2020review}.
Soft-robotics systems are an emerging paradigm that can enable multi-steering capabilities and complex stress-less interventions through narrow passageways. 
IPEI scenarios reflect an environment where the snake-like robot can use the wall as a support to propel forward. Bio-inspired robots imitate biological systems such as snake locomotion \cite{transeth2009survey, chen2015minimum}, octopus tentacles \cite{fras2018fluidical}, elephant trunks \cite{luo2017design}, and mammalian spine \cite{hu2019design}. They have been an emerging research direction in soft-robotic actuation \cite{kolachalama2020continuum}. Pressure-driven eversion of flexible, thin-walled tubes, called vine robots, has shown increased applications to navigate confined spaces \cite{hawkes2017soft}. 

\subsubsection{Proprioception and Shape-sensing} 
To achieve precise and reliable motion control of continuum robots, accurate and real-time shape sensing is needed. However, accurately modelling the robot shape is challenging due to friction, backlash, the inherent deformable nature of the lumen or vessels and inevitable collisions with the anatomy \cite{shi2016shape}. Some emerging sensor-based shape reconstruction techniques for interventional devices rely on \ac{FBG} and \ac{EM} sensors \cite{shi2016shape, sahu2021shape,ha2021robust, ha2022contact, ha2022shape}. 
Both \ac{FBG} and \ac{EM} enabled techniques provide real-time shape estimation due to their short response time,  miniature size, biocompatibility, non-toxicity, and high sensitivity. Multiple sensors can be attached along the length of the continuum robot to track the robot and measure the axial strain. However, \ac{FBG} sensors provide a poor response in high-strain conditions and \ac{EM} sensors suffer from the problem of \ac{EM} interference \cite{sahu2021shape}. Hence, a sensor-fusion method between \ac{FBG}, \ac{EM} sensors and sparse fluoroscopic images could improve 3D catheter shape reconstruction accuracy \cite{ha2022shape}.

\subsubsection{Lumen/vessel modelling} Intraoperative imaging modalities such as ultrasound and optical coherence tomography can support direct observation and visualisation \cite{barata2021ivus, zulina2021colon, liao2022distortion}. Sensor fusion between \ac{ivus} and \ac{EM} can provide an intravascular reconstruction of vessels \cite{shi2016real, barata2021ivus}. For computer-assisted navigation, \ac{SLAM} has been successfully demonstrated in inferring dense and detailed depth maps and lumen reconstruction \cite{chadebecq2020computer}. Depth prediction models have been developed recently to estimate lumen features \cite{rau2019implicit}.

\section{CONCLUSIONS}

Navigation is one of the crucial steps of IPEI that requires extensive interventional dexterity and skills. 
This work provides a detailed overview of several critical aspects required to improve IPEI navigation. We propose a classification of dedicated autonomy levels and provide a systematic review of the governing motion planning methods.
Autonomous navigation could improve the overall execution of IPEI procedures, enabling the interventionist to focus on the medical aspects rather than on control issues with the instruments.
Therefore, in this article, we define the levels of autonomy required for IPEI navigation and the foreseeable human intervention associated with each level. This classification will improve risk and safety management while we advance towards higher levels of autonomy.
One of the essential steps towards achieving automation is through employing MP methods. A comprehensive overview of MP techniques used in IPEI navigation is provided in this work. At the same time, the limitations associated with existing methods are provided.
These voids in capabilities need to be overcome if one wants to raise the level of autonomy of today's existing robotic systems.
These include improvements in MP techniques and in enhanced robotic capabilities such as actuation and proprioception modelling.
Autonomous navigation can positively impact IPEI procedures, making them widely accessible to a greater population.

\section{ACKNOWLEDGEMENTS}
This work was supported by the ATLAS project. This project has received funding from the European Union's Horizon 2020 research and innovation programme under the Marie Sklodowska-Curie grant agreement No 813782.

\begin{scriptsize}
\begin{acronym}[Orocos]
    \acro{ivus}[IVUS]{IntraVascular UltraSound}
    \acro{ACO}[ACO]{Ant Colony Optimization}
    \acro{BFS}[BFS]{Breadth First Search}
    \acro{BPH}[BPH]{Benign Prostatic Hyperplasia}
    \acro{CBS}[CBS]{Centerline-based Structure}
    \acro{CNN}[CNN]{Convolutional Neural Network}
    \acro{CRC}[CRC]{Colorectal Cancer}
    \acro{CRISP}[CRISP]{Continuum Reconfigurable Incisionless Surgical Parallel}
    \acro{CT}[CT]{Computed Tomography}
    \acro{CTRs}[CTRs]{Concentric Tube Robots}
    \acro{C-TBNA}[C-TBNA]{Conventional Transbronchial Needle Aspiration}
    \acro{DDPG}[DDPG]{Deep Deterministic policy gradient}
    \acro{DFS}[DFS]{Depth First Search}
    \acro{DMPs}[DMPs]{Dynamical Movement Primitives}
    \acro{DNN}[DNN]{Deep Neural Network}
    \acro{DoFs}[DoFs]{Degrees-of-Freedom}
    \acro{DQN}[DQN]{Deep Q-Network}
    \acro{DRL}[DRL]{Deep Reinforcement Learning}
    \acro{EBUS}[EBUS]{EndoBronchial UltraSound}
    \acro{EI}[EI]{Endovascular Interventions}
    \acro{ELSE}[ELSE]{Ethical, Legal, Social and Economic}
    \acro{EM}[EM]{Electromagnetic}
    \acro{ESD}[ESD]{Endoscopic Submucosal Disection}
    \acro{FBG}[FBG]{Fiber Bragg Gratings}
    \acro{FEM}[FEM]{Finite-Element Method}
    \acro{fURS}[fURS]{Robotic Flexible Ureteroscopy}
    \acro{GA}[GA]{Genetic Algorithm}
    \acro{GAN}[GAN]{Generative Adversarial Network}
    \acro{GAIL}[GAIL]{Generative Adversarial Imitation Learning}
    \acro{GMM}[GMM]{Gaussian Mixture Model}
    \acro{HMMs}[HMMs]{Hidden Markov models}
    \acro{IP}[IP]{Intraluminal Procedures}
    \acro{IPEI}[IPEI]{Intraluminal Procedures and Endovascular Interventions}
    \acro{LoA}[LoA]{Levels of Autonomy}
    \acro{LfD}[LfD]{Learning from Demonstrations}
    \acro{LPA*}[LPA*]{Lifelong Planning A*}
    \acro{L-BFGS}[L-BFGS]{Limited-memory Broyden-Fletcher-Goldfarb-Shanno}
    \acro{LSTM}[LSTM]{Long Short Term Memory}
    \acro{MIS}[MIS]{Minimally Invasive Surgical}
    \acro{MP}[MP]{Motion Planning}
    \acro{MRI}[MRI]{Magnetic Resonance Imaging}
    \acro{MSM}[MSM]{Mass-Spring Model}
    \acro{OCT}[OCT]{Optical Coherence Tomography}
    \acro{PBD}[PBD]{Position-Based Dynamics}
    \acro{PRM*}[PRM*]{Probabilistic RoadMap*}
    \acro{RL}[RL]{Reinforcement Learning}
    \acro{ROI}[ROI]{Region of Interest}
    \acro{RRG}[RRG]{Rapidly-exploring Random Graph}
    \acro{RRM}[RRM]{Rapidly-exploring RoadMap}
    \acro{RRT}[RRT]{Rapidly-exploring Random Tree}
    \acro{RRT-connect}[RRT-connect]{bi-directional RRT, or bi-RRT}
    \acro{SLAM}[SLAM]{Simultaneous Localisation And Mapping}
    \acro{TNE}[TNE]{Transnasal Endoscopy}
    \acro{TOE}[TOE]{Transoral Endoscopy}
    \acro{TORS}[TORS]{Transoral Robotic Surgery}
    \acro{TURBT}[TURBT]{Trans-Urethral Resection of Bladder Tumours}
    \acro{TURP}[TURP]{Trans-Urethral Resection of the Prostate}
\end{acronym}
\end{scriptsize}

\bibliographystyle{unsrt}
\bibliography{references}  

\begin{thebibliography}{100}

\bibitem{seetohul2022snake}
Jenna Seetohul et~al.
\newblock Snake robots for surgical applications: A review.
\newblock {\em Robot.}, 11(3):57, 2022.

\bibitem{da2020challenges}
Tomas da~Veiga et~al.
\newblock Challenges of continuum robots in clinical context: a review.
\newblock {\em Prog. Biomed. Eng.}, 2020.

\bibitem{simaan2018medical}
Nabil Simaan et~al.
\newblock Medical technologies and challenges of robot-assisted minimally
  invasive intervention and diagnostics.
\newblock {\em Annu. Rev. Control Robot. Auton. Syst.}, 1:465--490, 2018.

\bibitem{prendergast2018autonomous}
J~Micah Prendergast et~al.
\newblock Autonomous localization, navigation and haustral fold detection for
  robotic endoscopy.
\newblock In {\em Proc. IEEE Int. Conf. Intell. Robot. Syst.}, pages 783--790,
  2018.

\bibitem{hwang2020review}
Junsun Hwang et~al.
\newblock A review of magnetic actuation systems and magnetically actuated
  guidewire-and catheter-based microrobots for vascular interventions.
\newblock {\em Intell. Service Robot.}, 13(1):1--14, 2020.

\bibitem{orekhov2018snake}
A~Orekhov et~al.
\newblock Snake-like robots for minimally invasive, single-port, and
  intraluminal surgeries.
\newblock {\em The Enciclopedia of Med. Robot. World Sci.}, pages 203--243,
  2018.

\bibitem{manfredi2021endorobots}
Luigi Manfredi.
\newblock Endorobots for colonoscopy: Design challenges and available
  technologies.
\newblock {\em Front. in Robot. AI}, page 209, 2021.

\bibitem{fiorini2022concepts}
Paolo Fiorini et~al.
\newblock Concepts and trends in autonomy for robot-assisted surgery.
\newblock {\em Proc. IEEE}, 110(7):993--1011, 2022.

\bibitem{attanasio2020autonomy}
Aleks Attanasio et~al.
\newblock Autonomy in surgical robotics.
\newblock {\em Annu. Rev. Control Robot. Auton. Syst.}, 4, 2020.

\bibitem{haidegger2019autonomy}
Tam{\'a}s Haidegger.
\newblock Autonomy for surgical robots: Concepts and paradigms.
\newblock {\em IEEE Trans. Med. Robot. Bionics}, 1(2):65--76, 2019.

\bibitem{hargest2021five}
Rachel Hargest.
\newblock Five thousand years of minimal access surgery: 1990--present:
  organisational issues and the rise of the robots.
\newblock {\em J. Royal Soc. Medicine}, 114(2):69--76, 2021.

\bibitem{yang2017medical}
Guang-Zhong Yang et~al.
\newblock Medical robotics—regulatory, ethical, and legal considerations for
  increasing levels of autonomy.
\newblock {\em Sci. Robot.}, 2(4):8638, 2017.

\bibitem{patle2019review}
BK~Patle et~al.
\newblock A review: On path planning strategies for navigation of mobile robot.
\newblock {\em Defence Technol.}, 15(4):582--606, 2019.

\bibitem{latombe2012robot}
Jean-Claude Latombe.
\newblock {\em Robot motion planning}, volume 124.
\newblock Springer Science \& Business Media, 2012.

\bibitem{omisore2020review}
Olatunji~Mumini Omisore et~al.
\newblock A review on flexible robotic systems for minimally invasive surgery.
\newblock {\em IEEE Trans. Syst., Man, and Cybernetics: Syst.}, 2020.

\bibitem{burgner2015continuum}
Jessica Burgner-Kahrs et~al.
\newblock Continuum robots for medical applications: A survey.
\newblock {\em IEEE Trans. Robot.}, 31(6):1261--1280, 2015.

\bibitem{vitiello2012emerging}
Valentina Vitiello et~al.
\newblock Emerging robotic platforms for minimally invasive surgery.
\newblock {\em IEEE Rev. Biomed. Eng.}, 6:111--126, 2012.

\bibitem{blecha2020modern}
Matthew Blecha et~al.
\newblock Modern endovascular therapy.
\newblock {\em World J. Surg.}, pages 1--10, 2020.

\bibitem{villaret2017robotic}
Andrea~Bolzoni Villaret et~al.
\newblock Robotic transnasal endoscopic skull base surgery: systematic review
  of the literature and report of a novel prototype for a hybrid system
  (brescia endoscope assistant robotic holder).
\newblock {\em World neurosurgery}, 105:875--883, 2017.

\bibitem{peters2018review}
Brian~S Peters et~al.
\newblock Review of emerging surgical robotic technology.
\newblock {\em Surg. Endoscopy}, 32(4):1636--1655, 2018.

\bibitem{rafii2014current}
Hedyeh Rafii-Tari et~al.
\newblock Current and emerging robot-assisted endovascular catheterization
  technologies: a review.
\newblock {\em Ann. Biomed. Eng.}, 42(4):697--715, 2014.

\bibitem{bonatti2014robotic}
Johannes Bonatti et~al.
\newblock Robotic technology in cardiovascular medicine.
\newblock {\em Nature Rev. Cardiol.}, 11(5):266, 2014.

\bibitem{fu2009steerable}
Yili Fu et~al.
\newblock Steerable catheters in minimally invasive vascular surgery.
\newblock {\em Int. J. Med. Robot. Comput. Assist. Surg.}, 5(4):381--391, 2009.

\bibitem{pourdjabbar2017development}
Ali Pourdjabbar et~al.
\newblock The development of robotic technology in cardiac and vascular
  interventions.
\newblock {\em Rambam Maimonides Med. J.}, 8(3), 2017.

\bibitem{berczeli2021catheter}
Marton Berczeli et~al.
\newblock Catheter robots in the cardiovascular system.
\newblock {\em Latest Develop. Med. Robot. Syst.}, page~95, 2021.

\bibitem{ciuti2020frontiers}
Gastone Ciuti et~al.
\newblock Frontiers of robotic colonoscopy: A comprehensive review of robotic
  colonoscopes and technologies.
\newblock {\em J. Clinical Medicine}, 9(6):1648, 2020.

\bibitem{yeung2019emerging}
Chung-Kwong Yeung et~al.
\newblock Emerging next-generation robotic colonoscopy systems towards painless
  colonoscopy.
\newblock {\em J. Digestive Diseases}, 20(4):196--205, 2019.

\bibitem{you2019automatic}
Hyeonseok You et~al.
\newblock Automatic control of cardiac ablation catheter with deep
  reinforcement learning method.
\newblock {\em J. Mech. Sci. Tech.}, 33(11):5415--5423, 2019.

\bibitem{lemke2021colonoscopy}
Madeline Lemke et~al.
\newblock Colonoscopy trainers experience greater stress during insertion than
  withdrawal: implications for endoscopic curricula.
\newblock {\em J. Can. Assoc. Gastroenterology}, 4(1):15--20, 2021.

\bibitem{ahmed2019colonoscopy}
Rajib Ahmed et~al.
\newblock Colonoscopy technologies for diagnostics and drug delivery.
\newblock {\em Med. Devices \& Sensors}, 2(3-4):e10041, 2019.

\bibitem{wernli2016risks}
Karen~J Wernli et~al.
\newblock Risks associated with anesthesia services during colonoscopy.
\newblock {\em Gastroenterology}, 150(4):888--894, 2016.

\bibitem{hassan2019diagnostic}
Cesare Hassan et~al.
\newblock Diagnostic yield and miss rate of endorings in an organized
  colorectal cancer screening program: the smart (study methodology for
  adr-related technology) trial.
\newblock {\em Gastrointestinal Endoscopy}, 89(3):583--590, 2019.

\bibitem{than2015diagnostic}
Mary Than et~al.
\newblock Diagnostic miss rate for colorectal cancer: an audit.
\newblock {\em Ann. gastroenterology}, 28(1):94, 2015.

\bibitem{eickhoff2006vitro}
A~Eickhoff et~al.
\newblock In vitro evaluation of forces exerted by a new computer-assisted
  colonoscope (the neoguide endoscopy system).
\newblock {\em Endoscopy}, 38(12):1224--1229, 2006.

\bibitem{gluck2016novel}
Nathan Gluck et~al.
\newblock A novel self-propelled disposable colonoscope is effective for
  colonoscopy in humans (with video).
\newblock {\em Gastrointestinal endoscopy}, 83(5):998--1004, 2016.

\bibitem{shike2008sightline}
Moshe Shike et~al.
\newblock Sightline colonosight system for a disposable, power-assisted,
  non-fiber-optic colonoscopy (with video).
\newblock {\em Gastrointestinal endoscopy}, 68(4):701--710, 2008.

\bibitem{tumino2017use}
E~Tumino et~al.
\newblock Use of robotic colonoscopy in patients with previous incomplete
  colonoscopy.
\newblock {\em Eur. Rev. Med. Pharmacol Sci.}, 21(4):819--826, 2017.

\bibitem{herrell2014future}
S~Duke Herrell et~al.
\newblock Future robotic platforms in urologic surgery: recent developments.
\newblock {\em Current opinion in urology}, 24(1):118, 2014.

\bibitem{tyson2014urological}
Mark~D Tyson et~al.
\newblock Urological applications of natural orifice transluminal endoscopic
  surgery.
\newblock {\em Nature Rev. Urology}, 11(6):324--332, 2014.

\bibitem{bazzi2012natural}
Wassim~M Bazzi et~al.
\newblock Natural orifice transluminal endoscopic surgery in urology: Review of
  the world literature.
\newblock {\em Urology Ann.}, 4(1):1, 2012.

\bibitem{chen2020review}
Yuyang Chen et~al.
\newblock Review of surgical robotic systems for keyhole and endoscopic
  procedures: state of the art and perspectives.
\newblock {\em Front. of med.}, 14(4):382--403, 2020.

\bibitem{rassweiler2018robot}
Jens Rassweiler et~al.
\newblock Robot-assisted flexible ureteroscopy: an update.
\newblock {\em Urolithiasis}, 46(1):69--77, 2018.

\bibitem{gandaglia2016novel}
Giorgio Gandaglia et~al.
\newblock Novel technologies in urologic surgery: a rapidly changing scenario.
\newblock {\em Current urology Rep.}, 17(3):1--8, 2016.

\bibitem{valdastri2012advanced}
Pietro Valdastri et~al.
\newblock Advanced technologies for gastrointestinal endoscopy.
\newblock {\em Ann. Rev. Biomed. Eng.}, 14, 2012.

\bibitem{marlicz2020frontiers}
Wojciech Marlicz et~al.
\newblock Frontiers of robotic gastroscopy: A comprehensive review of robotic
  gastroscopes and technologies.
\newblock {\em Cancers}, 12(10):2775, 2020.

\bibitem{de2020trans}
Armando De~Virgilio et~al.
\newblock Trans-oral robotic surgery in the management of parapharyngeal space
  tumors: a systematic review.
\newblock {\em Oral Oncology}, 103:104581, 2020.

\bibitem{prakash1994bronchoscopy}
Udaya~BS Prakash et~al.
\newblock Bronchoscopy.
\newblock {\em J. Bronchology \& Interv. Pulmonology}, 1(4):340, 1994.

\bibitem{kennedy2020computer}
Lauren~R Kennedy-Metz et~al.
\newblock Computer vision in the operating room: Opportunities and caveats.
\newblock {\em IEEE Trans. Med. Robot. Bionics}, 3(1):2--10, 2020.

\bibitem{agrawal2020robotic}
Abhinav Agrawal et~al.
\newblock Robotic bronchoscopy for pulmonary lesions: a review of existing
  technologies and clinical data.
\newblock {\em J. thoracic disease}, 12(6):3279, 2020.

\bibitem{stammberger1990functional}
Heinz Stammberger et~al.
\newblock Functional endoscopic sinus surgery.
\newblock {\em Eur. archives of oto-rhino-laryngology}, 247(2):63--76, 1990.

\bibitem{burgner2013telerobotic}
Jessica Burgner et~al.
\newblock A telerobotic system for transnasal surgery.
\newblock {\em IEEE Trans. Mechatronics}, 19(3):996--1006, 2013.

\bibitem{madoglio2020robotics}
Alba Madoglio et~al.
\newblock Robotics in endoscopic transnasal skull base surgery: literature
  review and personal experience.
\newblock {\em Control Syst. Design of Bio-Robot. and Bio-mechatronics with
  Adv. Appl.}, pages 221--244, 2020.

\bibitem{tanuma2016current}
Tokuma Tanuma et~al.
\newblock Current status of transnasal endoscopy worldwide using ultrathin
  videoscope for upper gastrointestinal tract.
\newblock {\em Digestive Endoscopy}, 28:25--31, 2016.

\bibitem{simaan2009design}
Nabil Simaan et~al.
\newblock Design and integration of a telerobotic system for minimally invasive
  surgery of the throat.
\newblock {\em Int. J. Robot. Res.}, 28(9):1134--1153, 2009.

\bibitem{nof2009automation}
Shimon~Y Nof.
\newblock Automation: What it means to us around the world.
\newblock In {\em Springer Handbook Automat.}, pages 13--52. 2009.

\bibitem{olszewska2017ontology}
Joanna~Isabelle Olszewska et~al.
\newblock Ontology for autonomous robotics.
\newblock In {\em Proc. IEEE Int. Symp. Robot. Human Interact. Commun.}, pages
  189--194, 2017.

\bibitem{chen2021automation}
Hualong Chen et~al.
\newblock From automation system to autonomous system: An architecture
  perspective.
\newblock {\em J. Mar. Sci. Eng.}, 9(6):645, 2021.

\bibitem{fisher2021towards}
Michael Fisher et~al.
\newblock Towards a framework for certification of reliable autonomous systems.
\newblock {\em Auton. Agents and Multi-Agent Syst.}, 35(1):1--65, 2021.

\bibitem{ethics}
Federica Merenda et~al.
\newblock Ethics, safety and human centricity: Intelligent machines under the
  scope of the european ai regulation act.
\newblock In {\em Workshop Conf. Italian Inst. Robot. Intell. Mach.}, 2021.

\bibitem{o2019legal}
Shane O'sullivan et~al.
\newblock Legal, regulatory, and ethical frameworks for development of
  standards in artificial intelligence (ai) and autonomous robotic surgery.
\newblock {\em Int. J. Med. Robot. Comput. Assist. Surg.}, 15(1):e1968, 2019.

\bibitem{aiact}
European Parliament and Council of~the European~Union.
\newblock Artificial intelligence act: Regulation laying down harmonised rules
  on artificial intelligence and amending certain union legislative acts.
\newblock {\em Proposal for Regulation COM/2021/206}, 2021.

\bibitem{nationalaiact}
National artificial intelligence initiative.
\newblock
  \url{https://www.congress.gov/bill/116th-congress/house-bill/6216/text}.

\bibitem{fiazza2021icar}
Maria-Camilla Fiazza.
\newblock The eu proposal for regulating ai: Foreseeable impact on medical
  robotics.
\newblock In {\em Proc. IEEE Int. Conf. Adv. Robot.}, 2021.

\bibitem{medicaldevices}
International Electrotechnical~Commission (2017c).
\newblock Iec tr 60601-4-1 – medical electrical equipment – part 4-1:
  Guidance and interpretation - medical electrical equipment and medical
  electrical systems employing a degree of autonomy.
\newblock In {\em URL https://webstore.iec.ch/publication/29312.)}, 2017.

\bibitem{katic2015lapontospm}
Darko Kati{\'c} et~al.
\newblock Lapontospm: an ontology for laparoscopic surgeries and its
  application to surgical phase recognition.
\newblock {\em Int. J. Comput. Assist. Radiol. Surg.}, 10(9):1427--1434, 2015.

\bibitem{ravigopal2022fluoroscopic}
Sharan~R Ravigopal et~al.
\newblock Fluoroscopic image-based 3-d environment reconstruction and automated
  path planning for a robotically steerable guidewire.
\newblock {\em IEEE Robot. Automat. Lett.}, 7(4):11918--11925, 2022.

\bibitem{khare2015hands}
Rahul Khare et~al.
\newblock Hands-free system for bronchoscopy planning and guidance.
\newblock {\em IEEE Trans. Biomed. Eng.}, 62(12):2794--2811, 2015.

\bibitem{zheng20183d}
Jian-Qing Zheng et~al.
\newblock Towards 3d path planning from a single 2d fluoroscopic image for
  robot assisted fenestrated endovascular aortic repair.
\newblock In {\em Proc. IEEE Int. Conf. Robot. Automat.}, pages 8747--8753,
  2019.

\bibitem{zang2019optimal}
Xiaonan Zang et~al.
\newblock Optimal route planning for image-guided ebus bronchoscopy.
\newblock {\em Comput. in biology and medicine}, 112:103361, 2019.

\bibitem{taddese2018enhanced}
Addisu~Z Taddese et~al.
\newblock Enhanced real-time pose estimation for closed-loop robotic
  manipulation of magnetically actuated capsule endoscopes.
\newblock {\em Int. J. Robot. Res.}, 37(8):890--911, 2018.

\bibitem{page2021prisma}
Matthew~J Page et~al.
\newblock The prisma 2020 statement: an updated guideline for reporting
  systematic reviews.
\newblock {\em Brit. Med. J.}, 372, 2021.

\bibitem{rose2019pybliometrics}
Michael~E Rose et~al.
\newblock pybliometrics: Scriptable bibliometrics using a python interface to
  scopus.
\newblock {\em SoftwareX}, 10:100263, 2019.

\bibitem{vuik2021colon}
Fanny~ER Vuik et~al.
\newblock Colon capsule endoscopy in colorectal cancer screening: a systematic
  review.
\newblock {\em Endoscopy}, 53(08):815--824, 2021.

\bibitem{meng2019motion}
Ke~Meng et~al.
\newblock Motion planning and robust control for the endovascular navigation of
  a microrobot.
\newblock {\em IEEE Trans. Ind. Inform.}, 2019.

\bibitem{siciliano2008springer}
Bruno Siciliano et~al.
\newblock {\em Springer handbook of robotics}, volume 200.
\newblock Springer, 2008.

\bibitem{yang2016survey}
Liang Yang et~al.
\newblock Survey of robot 3d path planning algorithms.
\newblock {\em J. Control Sci. Eng.}, 2016.

\bibitem{cheng2012enhanced}
Irene Cheng et~al.
\newblock Enhanced segmentation and skeletonization for endovascular surgical
  planning.
\newblock In {\em Proc. Med. Imag.: Image-Guided Procedures, Robot. Interv.,
  and Model.}, volume 8316, pages 868--874, 2012.

\bibitem{tarjan1972depth}
Robert Tarjan.
\newblock Depth-first search and linear graph algorithms.
\newblock {\em SIAM J. Comput.}, 1(2):146--160, 1972.

\bibitem{dechter1985generalized}
Rina Dechter et~al.
\newblock Generalized best-first search strategies and the optimality of a.
\newblock {\em J. ACM}, 32(3):505--536, 1985.

\bibitem{dijkstra1959note}
Edsger~W Dijkstra et~al.
\newblock A note on two problems in connexion with graphs.
\newblock {\em Numerische mathematik}, 1(1):269--271, 1959.

\bibitem{hwang1992potential}
Yong~Koo Hwang et~al.
\newblock A potential field approach to path planning.
\newblock {\em IEEE Trans. Robot. Automat.}, 8(1):23--32, 1992.

\bibitem{hart1968formal}
Peter~E Hart et~al.
\newblock A formal basis for the heuristic determination of minimum cost paths.
\newblock {\em IEEE Trans. Syst. Sci. Cybernetics}, 4(2):100--107, 1968.

\bibitem{lavalle1998rapidly}
Steven~M LaValle et~al.
\newblock Rapidly-exploring random trees: A new tool for path planning.
\newblock {\em Comput. Sci. Dept. Oct.}, 98(11), 1998.

\bibitem{geraerts2004comparative}
Roland Geraerts et~al.
\newblock A comparative study of probabilistic roadmap planners.
\newblock In {\em Algorithmic Found. Robot.}, pages 43--57. 2004.

\bibitem{karaman2011sampling}
Sertac Karaman et~al.
\newblock Sampling-based algorithms for optimal motion planning.
\newblock {\em Int. J. Robot. Res.}, 30(7):846--894, 2011.

\bibitem{raja2012optimal}
Purushothaman Raja et~al.
\newblock Optimal path planning of mobile robots: A review.
\newblock {\em Int. J. Phys. Sci.}, 7(9):1314--1320, 2012.

\bibitem{dorigo2006ant}
Marco Dorigo et~al.
\newblock Ant colony optimization.
\newblock {\em IEEE Comput. Intell. Mag.}, 1(4):28--39, 2006.

\bibitem{ravichandar2020recent}
Harish Ravichandar et~al.
\newblock Recent advances in robot learning from demonstration.
\newblock {\em Annu. Rev. Control, Robot., and Auton. Syst.}, 3:297--330, 2020.

\bibitem{sutton2018reinforcement}
Richard~S Sutton et~al.
\newblock {\em Reinforcement learning: An introduction}.
\newblock MIT press, 2018.

\bibitem{Geiger_2005}
Bernhard Geiger et~al.
\newblock Virtual bronchoscopy of peripheral nodules using arteries as
  surrogate pathways.
\newblock In Amir~A. Amini and Armando Manduca, editors, {\em Proc. Med. Imag.:
  Physiol. Func. Struct. from Med. Imag.} SPIE, 2005.

\bibitem{S_nchez_2016}
Carles S{\'{a}}nchez et~al.
\newblock Navigation path retrieval from videobronchoscopy using bronchial
  branches.
\newblock In {\em Workshop on Clin. Image. Based Procedures}, pages 62--70.
  2016.

\bibitem{Wang_2011}
Junchen Wang et~al.
\newblock Intravascular catheter navigation using path planning and virtual
  visual feedback for oral cancer treatment.
\newblock {\em Int. J. Med. Robot. Comput. Assist. Surg.}, 7(2):214--224, 2011.

\bibitem{yang2014centerlines}
Fan Yang et~al.
\newblock Centerlines extraction for lumen model of human vasculature for
  computer-aided simulation of intravascular procedures.
\newblock In {\em Proc. IEEE World Congr. Intell. Control Automat.}, pages
  970--975, 2014.

\bibitem{kerschnitzki2013architecture}
Michael Kerschnitzki et~al.
\newblock Architecture of the osteocyte network correlates with bone material
  quality.
\newblock {\em J. bone and mineral Res.}, 28(8):1837--1845, 2013.

\bibitem{yudong2021rapid}
WANG Yudong et~al.
\newblock Rapid path extraction and three-dimensional roaming of the virtual
  endonasal endoscope.
\newblock {\em Chin. J. Electronics}, 30(3):397--405, 2021.

\bibitem{zang2021image}
Xiaonan Zang et~al.
\newblock Image-guided ebus bronchoscopy system for lung-cancer staging.
\newblock {\em Inform. in medicine unlocked}, 25:100665, 2021.

\bibitem{gibbs2013optimal}
Jason~D Gibbs et~al.
\newblock Optimal procedure planning and guidance system for peripheral
  bronchoscopy.
\newblock {\em IEEE Trans. Biomed. Eng.}, 61(3):638--657, 2013.

\bibitem{huang2011interactive}
Dongjin Huang et~al.
\newblock An interactive 3d preoperative planning and training system for
  minimally invasive vascular surgery.
\newblock In {\em Pro. IEEE Int. Conf. Comput. Aided Des. Comput. Graph.},
  pages 443--449, 2011.

\bibitem{fischer2022using}
Cedric Fischer et~al.
\newblock Using magnetic fields to navigate and simultaneously localize
  catheters in endoluminal environments.
\newblock {\em IEEE Robot. Automat. Lett.}, 2022.

\bibitem{Schafer_2007}
Sebastian Schafer et~al.
\newblock Planning image-guided endovascular interventions: guidewire
  simulation using shortest path algorithms.
\newblock In {\em Proc. Med. Imag.: Visualization and Imag. Guided Procedures},
  volume 6509, pages 813--822, 2007.

\bibitem{egger2007}
J.~{Egger} et~al.
\newblock A fast vessel centerline extraction algorithm for catheter
  simulation.
\newblock In {\em Proc. IEEE Int. Symp. Comput. Based Med. Syst.}, pages
  177--182, 2007.

\bibitem{liu2010vitro}
H~Liu et~al.
\newblock An in vitro investigation of image-guided steerable catheter
  navigation.
\newblock {\em Proc. Institution of Mech. Eng.: J. Eng. Medicine},
  224(8):945--954, 2010.

\bibitem{gibbs20073d}
Jason~D Gibbs et~al.
\newblock 3d path planning and extension for endoscopic guidance.
\newblock In {\em Med. Imag.: Visualization and Imag. Guided Procedures}, 2007.

\bibitem{gibbs2008integrated}
Jason~D Gibbs et~al.
\newblock Integrated system for planning peripheral bronchoscopic procedures.
\newblock In {\em Med. Imag.: Physiol. Func. Struct. Med. Imag.}, 2008.

\bibitem{qian2019towards}
Hanxin Qian et~al.
\newblock Towards rebuild the interventionist's intra-operative natural
  behavior: A fully sensorized endovascular robotic system design.
\newblock In {\em Proc. IEEE Int. Conf. Med. Imag. Phys. Eng.}, 2019.

\bibitem{schegg2022automated}
Pierre Schegg et~al.
\newblock Automated planning for robotic guidewire navigation in the coronary
  arteries.
\newblock In {\em Proc. IEEE Int. Conf. Soft Robot.}, pages 239--246, 2022.

\bibitem{cho2021image}
Yongjun Cho et~al.
\newblock Image processing based autonomous guidewire navigation in
  percutaneous coronary intervention.
\newblock In {\em Proc. IEEE Int. Conf. Consum. Electron. Asia}, 2021.

\bibitem{rosell2012motion}
Jan Rosell et~al.
\newblock Motion planning for the virtual bronchoscopy.
\newblock In {\em Proc. IEEE Int. Conf. Robot. Automat.}, pages 2932--2937,
  2012.

\bibitem{yang2019path}
Fan Yang et~al.
\newblock Path planning of flexible ureteroscope based on ct image.
\newblock In {\em Proc. IEEE Chin. Control Conf.}, pages 4667--4672, 2019.

\bibitem{martin2020enabling}
James~W Martin et~al.
\newblock Enabling the future of colonoscopy with intelligent and autonomous
  magnetic manipulation.
\newblock {\em Nature Mach. Intell.}, 2(10):595--606, 2020.

\bibitem{zhang2020enabling}
Qi~Zhang et~al.
\newblock Enabling autonomous colonoscopy intervention using a robotic
  endoscope platform.
\newblock {\em IEEE Trans. Biomed. Eng.}, 68(6):1957--1968, 2020.

\bibitem{girerd2020slam}
Cedric Girerd et~al.
\newblock Slam-based follow-the-leader deployment of concentric tube robots.
\newblock {\em IEEE Robot. Automat. Lett.}, 5(2):548--555, 2020.

\bibitem{he2020}
Yucheng He et~al.
\newblock Endoscopic path planning in robot-assisted endoscopic nasal surgery.
\newblock {\em IEEE Access}, 2020.

\bibitem{ciobirca2018new}
C~Ciobirca et~al.
\newblock A new procedure for automatic path planning in bronchoscopy.
\newblock {\em Mater. Today: Proc.}, 5(13):26513--26518, 2018.

\bibitem{niyaz2018following}
Sherdil Niyaz et~al.
\newblock Following surgical trajectories with concentric tube robots via
  nearest-neighbor graphs.
\newblock In {\em Proc. Int. Symp. Exp. Robot.}, 2018.

\bibitem{niyaz2019optimizing}
Sherdil Niyaz et~al.
\newblock Optimizing motion-planning problem setup via bounded evaluation with
  application to following surgical trajectories.
\newblock In {\em Proc. IEEE Int. Conf. Intell. Robot. Syst.}, pages
  1355--1362, 2019.

\bibitem{koenig2004design}
Nathan Koenig et~al.
\newblock Design and use paradigms for gazebo, an open-source multi-robot
  simulator.
\newblock In {\em Proc. IEEE Int. Conf. Intell. Robot. Syst.}, volume~3, pages
  2149--2154, 2004.

\bibitem{ravigopal2021automated}
Sharan~R Ravigopal et~al.
\newblock Automated motion control of the coast robotic guidewire under
  fluoroscopic guidance.
\newblock In {\em Proc. Int. Symp. Med. Robot.}, 2021.

\bibitem{huang2021autonomous}
Hao-En Huang et~al.
\newblock Autonomous navigation of a magnetic colonoscope using force sensing
  and a heuristic search algorithm.
\newblock {\em Sci. Rep.}, 11(1):1--15, 2021.

\bibitem{fagogenis2019autonomous}
G~Fagogenis et~al.
\newblock Autonomous robotic intracardiac catheter navigation using haptic
  vision.
\newblock {\em Sci. Robot.}, 4(29), 2019.

\bibitem{aguilar2017rrt}
Wilbert~G Aguilar et~al.
\newblock Rrt-based path planning for virtual bronchoscopy simulator.
\newblock In {\em Proc. IEEE Int. Conf. Augmented Reality, Virtual Reality and
  Comput. Graph.}, pages 155--165, 2017.

\bibitem{aguilar2017virtual}
Wilbert~G Aguilar et~al.
\newblock Virtual bronchoscopy motion planner.
\newblock In {\em Proc. IEEE Int. Conf. Electronics, Electrical Eng. and
  Comput.}, 2017.

\bibitem{fellmann2015implications}
Carolin Fellmann et~al.
\newblock Implications of trajectory generation strategies for tubular
  continuum robots.
\newblock In {\em Proc. IEEE Int. Conf. Intell. Robot. Syst.}, pages 202--208,
  2015.

\bibitem{kuntz2015motion}
Alan Kuntz et~al.
\newblock Motion planning for a three-stage multilumen transoral lung access
  system.
\newblock In {\em Proc. IEEE Int. Conf. Intell. Robot. Syst.}, pages
  3255--3261, 2015.

\bibitem{guo2021training}
Jian Guo et~al.
\newblock A training system for vascular interventional surgeons based on local
  path planning.
\newblock In {\em Proc. IEEE Int. Conf. Mechatronics and Automat.}, pages
  1328--1333, 2021.

\bibitem{alterovitz2011rapidly}
Ron Alterovitz et~al.
\newblock Rapidly-exploring roadmaps: Weighing exploration vs. refinement in
  optimal motion planning.
\newblock In {\em Proc. IEEE Int. Conf. Robot. Automat.}, pages 3706--3712,
  2011.

\bibitem{torres2011motion}
Luis~G. Torres et~al.
\newblock Motion planning for concentric tube robots using mechanics-based
  models.
\newblock In {\em Proc. IEEE Int. Conf. Intell. Robot. Syst.}, 2011.

\bibitem{torres2012task}
Luis~G Torres et~al.
\newblock Task-oriented design of concentric tube robots using mechanics-based
  models.
\newblock In {\em Proc. IEEE Int. Conf. Intell. Robot. Syst.}, pages
  4449--4455, 2012.

\bibitem{torres2014interactive}
Luis~G Torres et~al.
\newblock Interactive-rate motion planning for concentric tube robots.
\newblock In {\em Proc. IEEE Int. Conf. Robot. Automat.}, pages 1915--1921,
  2014.

\bibitem{fauser2018planning}
Johannes Fauser et~al.
\newblock Planning nonlinear access paths for temporal bone surgery.
\newblock {\em Int. J. Comput. Assist. Radiol. Surg.}, 13(5):637--646, 2018.

\bibitem{fauser2018generalized}
Johannes Fauser et~al.
\newblock Generalized trajectory planning for nonlinear interventions.
\newblock In {\em Comput. Assist. Robot. Endoscopy}. pp. 46--53, 2018.

\bibitem{fauser2019optimizing}
Johannes Fauser et~al.
\newblock Optimizing clearance of b{\'e}zier spline trajectories for
  minimally-invasive surgery.
\newblock In {\em Proc. Int. Conf. Med. Imag. Comput. and Imag. Assist.
  Interv.}, pages 20--28, 2019.

\bibitem{fauser2019planning}
Johannes Fauser et~al.
\newblock Planning for flexible surgical robots via b{\'e}zier spline
  translation.
\newblock {\em IEEE Robot. Automat. Lett.}, 4(4):3270--3277, 2019.

\bibitem{kuntz2019planning}
Alan Kuntz et~al.
\newblock Planning high-quality motions for concentric tube robots in point
  clouds via parallel sampling and optimization.
\newblock In {\em Proc. IEEE Int. Conf. Intell. Robot. Syst.}, pages
  2205--2212, 2019.

\bibitem{lyons2010planning}
Lisa~A Lyons et~al.
\newblock Planning active cannula configurations through tubular anatomy.
\newblock In {\em Proc. IEEE Int. Conf. Robot. Automat.}, pages 2082--2087,
  2010.

\bibitem{liu1989limited}
Dong~C Liu et~al.
\newblock On the limited memory bfgs method for large scale optimization.
\newblock {\em Math. Program.}, 45(1):503--528, 1989.

\bibitem{bazaraa2013nonlinear}
Mokhtar~S Bazaraa et~al.
\newblock {\em Nonlinear programming: theory and algorithms}.
\newblock John Wiley \& Sons, 2013.

\bibitem{gao2015three}
Mingke Gao et~al.
\newblock Three-dimensional path planning and guidance of leg vascular based on
  improved ant colony algorithm in augmented reality.
\newblock {\em J. Med. Syst.}, 39(11):133, 2015.

\bibitem{qi2019kinematic}
Fei Qi et~al.
\newblock Kinematic analysis and navigation method of a cable-driven continuum
  robot used for minimally invasive surgery.
\newblock {\em Int. J. Med. Robot. Comput. Assist. Surg.}, 15(4):e2007, 2019.

\bibitem{li2021path}
Zhen Li et~al.
\newblock Path planning for endovascular catheterization under curvature
  constraints via two-phase searching approach.
\newblock {\em Int. J. Comput. Assist. Radiol. Surg.}, 16(4):619--627, 2021.

\bibitem{guo2021design}
Jian Guo et~al.
\newblock Design a novel of path planning method for the vascular
  interventional surgery robot based on dwa model.
\newblock In {\em Proc. IEEE Int. Conf. Mechatronics and Automat.}, pages
  1322--1327, 2021.

\bibitem{abah2021image}
Colette Abah et~al.
\newblock Image-guided optimization of robotic catheters for patient-specific
  endovascular intervention.
\newblock In {\em Proc. IEEE Int. Symp. Med. Robot.}, pages 1--8, 2021.

\bibitem{zhao2022surgical}
Yan Zhao et~al.
\newblock Surgical gan: Towards real-time path planning for passive flexible
  tools in endovascular surgeries.
\newblock {\em Neurocomputing}, 2022.

\bibitem{meng2021evaluation}
Fanxu Meng et~al.
\newblock Evaluation of a reinforcement learning algorithm for vascular
  intervention surgery.
\newblock In {\em Proc. IEEE Int. Conf. Mechatronics and Automat.}, pages
  1033--1037, 2021.

\bibitem{trovato2010development}
Gabriele Trovato et~al.
\newblock Development of a colon endoscope robot that adjusts its locomotion
  through the use of reinforcement learning.
\newblock {\em Int. J. Comput. Assist. Radiol. Surg.}, 5(4):317--325, 2010.

\bibitem{Rafii_Tari_2013}
Hedyeh Rafii-Tari et~al.
\newblock Learning-based modeling of endovascular navigation for collaborative
  robotic catheterization.
\newblock In {\em Adv. Inf. Syst. Eng.}, pages 369--377. 2013.

\bibitem{rafii2014hierarchical}
Hedyeh Rafii-Tari et~al.
\newblock Hierarchical hmm based learning of navigation primitives for
  cooperative robotic endovascular catheterization.
\newblock In {\em Proc. Int. Conf. Med. Imag. Comput. and Comput. Assist.
  Interv.}, pages 496--503, 2014.

\bibitem{Chi_2018}
Wenqiang Chi et~al.
\newblock Trajectory optimization of robot-assisted endovascular
  catheterization with reinforcement learning.
\newblock In {\em Proc. IEEE Int. Conf. Intell. Robot. Syst.}, 2018.

\bibitem{Chi_2018_2}
Wenqiang Chi et~al.
\newblock Learning-based endovascular navigation through the use of non-rigid
  registration for collaborative robotic catheterization.
\newblock {\em Int. J. Comput. Assist. Radiol. Surg.}, 13(6):855--864, 2018.

\bibitem{chi_2020}
Wenqiang Chi et~al.
\newblock Collaborative robot-assisted endovascular catheterization with
  generative adversarial imitation learning.
\newblock In {\em Proc. IEEE Int. Conf. Robot. Automat.}, 2020.

\bibitem{behr2019deep}
Tobias Behr et~al.
\newblock Deep reinforcement learning for the navigation of neurovascular
  catheters.
\newblock {\em Current Directions Biomed. Eng.}, 5(1):5--8, 2019.

\bibitem{karstensen2020autonomous}
Lennart Karstensen et~al.
\newblock Autonomous guidewire navigation in a two dimensional vascular
  phantom.
\newblock {\em Current Directions in Biomed. Eng.}, 6(1), 2020.

\bibitem{kweon2021deep}
Jihoon Kweon et~al.
\newblock Deep reinforcement learning for guidewire navigation in coronary
  artery phantom.
\newblock {\em IEEE Access}, 9:166409--166422, 2021.

\bibitem{pore2022colonoscopy}
Ameya Pore et~al.
\newblock Colonoscopy navigation using end-to-end deep visuomotor control: A
  user study.
\newblock {\em arXiv preprint arXiv:2206.15086}, 2022.

\bibitem{karstensen2022learning}
Lennart Karstensen et~al.
\newblock Learning-based autonomous vascular guidewire navigation without human
  demonstration in the venous system of a porcine liver.
\newblock {\em Int. J. Comput. Assist. Radiol. Surg.}, pages 1--8, 2022.

\bibitem{saveriano2021dynamic}
Matteo Saveriano et~al.
\newblock Dynamic movement primitives in robotics: A tutorial survey.
\newblock {\em arXiv preprint arXiv:2102.03861}, 2021.

\bibitem{mnih2015human}
Volodymyr Mnih et~al.
\newblock Human-level control through deep reinforcement learning.
\newblock {\em Nature}, 518(7540):529--533, 2015.

\bibitem{athiniotisdeep}
S~Athiniotis et~al.
\newblock Deep q reinforcement learning for autonomous navigation of surgical
  snake robot in confined spaces.
\newblock In {\em Proc. Hamlyn Symp. Med. Robot.}, 2019.

\bibitem{ibarz2021train}
Julian Ibarz et~al.
\newblock How to train your robot with deep reinforcement learning: lessons we
  have learned.
\newblock {\em Int. J. Robot. Res.}, 40(4-5):698--721, 2021.

\bibitem{garcia2015comprehensive}
Javier Garc{\i}a et~al.
\newblock A comprehensive survey on safe reinforcement learning.
\newblock {\em J. Mach. Learn. Res.}, 16(1):1437--1480, 2015.

\bibitem{pore2021safe}
Ameya Pore et~al.
\newblock Safe reinforcement learning using formal verification for tissue
  retraction in autonomous robotic-assisted surgery.
\newblock In {\em Proc. IEEE Int. Conf. Intell. Robot. Syst.}, pages
  4025--4031. IEEE, 2021.

\bibitem{corsi2023constrained}
Davide Corsi et~al.
\newblock Constrained reinforcement learning and formal verification for safe
  colonoscopy navigation.
\newblock {\em arXiv preprint arXiv:2303.03207}, 2023.

\bibitem{lecun2015deep}
Yann LeCun et~al.
\newblock Deep learning.
\newblock {\em Nature}, 521(7553):436--444, 2015.

\bibitem{birkhoff2021review}
David~C Birkhoff et~al.
\newblock A review on the current applications of artificial intelligence in
  the operating room.
\newblock {\em Surg. Innov.}, 28(5):611--619, 2021.

\bibitem{goldenberg2017using}
Mitchell~G Goldenberg et~al.
\newblock Using data to enhance performance and improve quality and safety in
  surgery.
\newblock {\em JAMA Surg.}, 152(10):972--973, 2017.

\bibitem{rusu2016sim}
Andrei~A Rusu et~al.
\newblock Sim-to-real robot learning from pixels with progressive nets.
\newblock {\em arXiv preprint arXiv:1610.04286}, 2016.

\bibitem{schulman2017proximal}
John Schulman et~al.
\newblock Proximal policy optimization algorithms.
\newblock {\em arXiv preprint arXiv:1707.06347}, 2017.

\bibitem{haarnoja2018soft}
Tuomas Haarnoja et~al.
\newblock Soft actor-critic algorithms and applications.
\newblock {\em arXiv preprint arXiv:1812.05905}, 2018.

\bibitem{lin2020softgym}
Xingyu Lin et~al.
\newblock Softgym: Benchmarking deep reinforcement learning for deformable
  object manipulation.
\newblock {\em arXiv preprint arXiv:2011.07215}, 2020.

\bibitem{li2022position}
Zhen Li et~al.
\newblock Position-based dynamics simulator of vessel deformations for path
  planning in robotic endovascular catheterization.
\newblock {\em Med. Eng. \& Physics}, 110:103920, 2022.

\bibitem{dupont2022continuum}
Pierre Dupont et~al.
\newblock Continuum robots for medical interventions.
\newblock {\em Proc. IEEE}, 2022.

\bibitem{pore2021learning}
Ameya Pore et~al.
\newblock Learning from demonstrations for autonomous soft-tissue retraction.
\newblock In {\em Proc. IEEE Int. Symp. Med. Robot.}, 2021.

\bibitem{ghasemipour2020divergence}
Seyed Kamyar~Seyed Ghasemipour et~al.
\newblock A divergence minimization perspective on imitation learning methods.
\newblock In {\em Proc. PMLR Conf. Robot. Learn.}, pages 1259--1277, 2020.

\bibitem{liu2016fast}
Fangde Liu et~al.
\newblock Fast and adaptive fractal tree-based path planning for programmable
  bevel tip steerable needles.
\newblock {\em IEEE Robot. Automat. Lett.}, 1(2):601--608, 2016.

\bibitem{pinzi2019adaptive}
Marlene Pinzi et~al.
\newblock The adaptive hermite fractal tree (ahft): a novel surgical 3d path
  planning approach with curvature and heading constraints.
\newblock {\em Int. J. Comput. Assist. Radiol. Surg.}, 14(4):659--670, 2019.

\bibitem{wang2018learning}
Wei Wang et~al.
\newblock A learning-based multi-rrt approach for robot path planning in narrow
  passages.
\newblock {\em J. Intell. Robot. Syst.}, 90(1-2):81--100, 2018.

\bibitem{granna2019computer}
Josephine Granna et~al.
\newblock Computer-assisted planning for a concentric tube robotic system in
  neurosurgery.
\newblock {\em Int. J. Comput. Assist. Radiol. Surg.}, 14(2):335--344, 2019.

\bibitem{pourmanda2019navigation}
Mohammad~Javad Pourmanda et~al.
\newblock Navigation and control of endovascular helical swimming microrobot
  using dynamic programing and adaptive sliding mode strategy.
\newblock {\em Control Syst. Des. Bio-Robot. Bio-Mechatronics Adv. Appl.}, page
  201, 2019.

\bibitem{howell2019altro}
Taylor~A Howell et~al.
\newblock Altro: A fast solver for constrained trajectory optimization.
\newblock In {\em Proc. IEEE Int. Conf. Intell. Robot. Syst.}, pages
  7674--7679, 2019.

\bibitem{cheng2019end}
Richard Cheng et~al.
\newblock End-to-end safe reinforcement learning through barrier functions for
  safety-critical continuous control tasks.
\newblock In {\em Proc. AAAI Conf. Artif. Intell.}, volume~33, pages
  3387--3395, 2019.

\bibitem{prudencio2022survey}
Rafael~Figueiredo Prudencio et~al.
\newblock A survey on offline reinforcement learning: Taxonomy, review, and
  open problems.
\newblock {\em arXiv preprint arXiv:2203.01387}, 2022.

\bibitem{you2017model}
Xuanke You et~al.
\newblock Model-free control for soft manipulators based on reinforcement
  learning.
\newblock In {\em Proc. IEEE Int. Conf. Intell. Robot. Syst.}, pages
  2909--2915, 2017.

\bibitem{thuruthel2018model}
Thomas~George Thuruthel et~al.
\newblock Model-based reinforcement learning for closed-loop dynamic control of
  soft robotic manipulators.
\newblock {\em IEEE Trans. Robot.}, 35(1):124--134, 2018.

\bibitem{liu2020efficient}
Junjia Liu et~al.
\newblock Efficient reinforcement learning control for continuum robots based
  on inexplicit prior knowledge.
\newblock {\em arXiv preprint arXiv:2002.11573}, 2020.

\bibitem{pore2020simple}
Ameya Pore et~al.
\newblock On simple reactive neural networks for behaviour-based reinforcement
  learning.
\newblock {\em Proc. IEEE Int. Conf. Robot. Automat.}, 2020.

\bibitem{bengio2009curriculum}
Yoshua Bengio et~al.
\newblock Curriculum learning.
\newblock In {\em Proc. Annu. Int. Conf. Mach. Learn.}, pages 41--48, 2009.

\bibitem{transeth2009survey}
Aksel~Andreas Transeth et~al.
\newblock A survey on snake robot modeling and locomotion.
\newblock {\em Robot.}, 27(7):999--1015, 2009.

\bibitem{chen2015minimum}
Yanjie Chen et~al.
\newblock Minimum sweeping area motion planning for flexible serpentine
  surgical manipulator with kinematic constraints.
\newblock In {\em Proc. IEEE Int. Conf. Intell. Robot. Syst.}, pages
  6348--6353, 2015.

\bibitem{fras2018fluidical}
Jan Fras et~al.
\newblock Fluidical bending actuator designed for soft octopus robot tentacle.
\newblock In {\em Proc. IEEE Int. Conf. Soft Robot.}, pages 253--257, 2018.

\bibitem{luo2017design}
Ruidong Luo et~al.
\newblock Design and kinematic analysis of an elephant-trunk-like robot with
  shape memory alloy actuators.
\newblock In {\em Proc. IEEE Adv. Inf. Technol., Electronic and Automat.
  Control Conf.}, pages 157--161, 2017.

\bibitem{hu2019design}
Yang Hu et~al.
\newblock Design and fabrication of a 3-d printed metallic flexible joint for
  snake-like surgical robot.
\newblock {\em IEEE Robot. Automat. Lett.}, 4(2):1557--1563, 2019.

\bibitem{kolachalama2020continuum}
Srikanth Kolachalama et~al.
\newblock Continuum robots for manipulation applications: A survey.
\newblock {\em J. Robot.}, 2020.

\bibitem{hawkes2017soft}
Elliot~W Hawkes et~al.
\newblock A soft robot that navigates its environment through growth.
\newblock {\em Sci. Robot.}, 2(8), 2017.

\bibitem{shi2016shape}
Chaoyang Shi et~al.
\newblock Shape sensing techniques for continuum robots in minimally invasive
  surgery: A survey.
\newblock {\em IEEE Trans. Biomed. Eng.}, 64(8):1665--1678, 2016.

\bibitem{sahu2021shape}
Sujit~Kumar Sahu et~al.
\newblock Shape reconstruction processes for interventional application
  devices: State of the art, progress, and future directions.
\newblock {\em Front. in Robot. and AI}, 8:758411, 2021.

\bibitem{ha2021robust}
Xuan~Thao Ha et~al.
\newblock Robust catheter tracking by fusing electromagnetic tracking, fiber
  bragg grating and sparse fluoroscopic images.
\newblock {\em IEEE Sensors J.}, 21(20):23422--23434, 2021.

\bibitem{ha2022contact}
Xuan~Thao Ha et~al.
\newblock Contact localization of continuum and flexible robot using
  data-driven approach.
\newblock {\em IEEE Robot. Automat. Lett.}, 2022.

\bibitem{ha2022shape}
Xuan~Thao Ha, , et~al.
\newblock Shape sensing of flexible robots based on deep learning.
\newblock {\em IEEE Trans. Robot.}, 2022.

\bibitem{barata2021ivus}
Beatriz~Farola Barata et~al.
\newblock Ivus-based local vessel estimation for robotic intravascular
  navigation.
\newblock {\em IEEE Robot. Automat. Lett.}, 6(4):8102--8109, 2021.

\bibitem{zulina2021colon}
Natalia Zulina et~al.
\newblock Colon phantoms with cancer lesions for endoscopic characterization
  with optical coherence tomography.
\newblock {\em Biomed. Optics Express}, 12(2):955--968, 2021.

\bibitem{liao2022distortion}
Guiqiu Liao et~al.
\newblock Distortion and instability compensation with deep learning for
  rotational scanning endoscopic optical coherence tomography.
\newblock {\em Med. Image Anal.}, 77:102355, 2022.

\bibitem{shi2016real}
Chaoyang Shi et~al.
\newblock Real-time in vitro intravascular reconstruction and navigation for
  endovascular aortic stent grafting.
\newblock {\em Int. J. Med. Robot. and Comput. Assist. Surg.}, 12(4):648--657,
  2016.

\bibitem{chadebecq2020computer}
Fran{\c{c}}ois Chadebecq et~al.
\newblock Computer vision in the surgical operating room.
\newblock {\em Visceral Medicine}, 36(6):456--462, 2020.

\bibitem{rau2019implicit}
Anita Rau et~al.
\newblock Implicit domain adaptation with conditional generative adversarial
  networks for depth prediction in endoscopy.
\newblock {\em Int. J. Comput. Assist. Radiol. Surg.}, 14(7):1167--1176, 2019.

\end{thebibliography}

\end{document}